\documentclass{article} 
\usepackage{iclr2027_conference,times}

\usepackage{amsmath,amsfonts,bm}

\def\eqref#1{equation~\ref{#1}}

\def\1{\bm{1}}

\DeclareMathAlphabet{\mathsfit}{\encodingdefault}{\sfdefault}{m}{sl}
\SetMathAlphabet{\mathsfit}{bold}{\encodingdefault}{\sfdefault}{bx}{n}

\usepackage{hyperref}
\usepackage{url}
\usepackage[utf8]{inputenc}
\usepackage[english]{babel}
\usepackage{amssymb}
\usepackage{graphicx}
\usepackage{booktabs,multirow}
\usepackage{algorithm}
\usepackage{algpseudocode}
\usepackage{microtype}
\usepackage{xcolor}

\title{LD-EnFF: Latent-Dynamics Ensemble Flow Filtering for Data Assimilation with Sparse Observations}

\iclrfinalcopy

\author{Ziyu Tian\textsuperscript{1}, Kaichen Shen\textsuperscript{1}, Wenbo Hao\textsuperscript{1},
Phillip Si\textsuperscript{2}, Peng Chen\textsuperscript{2$\dagger$}, Wei Zhu\textsuperscript{1$\dagger$}\\[2pt]
\small \textsuperscript{1}School of Mathematics, Georgia Institute of Technology\\
\small \textsuperscript{2}School of Computational Science and Engineering, Georgia Institute of Technology\\
\textsuperscript{$\dagger$}Corresponding authors: pchen402@gatech.edu; weizhu@gatech.edu}

\begin{document}

\maketitle

\lhead{Preprint}

\begin{abstract}
Data assimilation combines model forecasts with noisy, incomplete observations to estimate the evolving state of a dynamical system. Existing methods face two compounding challenges: high-dimensional nonlinear dynamics make repeated forward simulation computationally expensive, while sparse observations provide limited direct information about the full state. To address these challenges, we propose the Latent-Dynamics Ensemble Flow Filter (LD-EnFF), a sequential Bayesian filtering framework that performs both forecast propagation and filtering updates in a compact latent space. LD-EnFF combines a latent dynamics surrogate for ensemble propagation with a variational autoencoder (VAE)-based observation model that evaluates a state-dependent observation likelihood in latent space. At each assimilation step, an ensemble filtering update based on flow matching uses the forecast ensemble and this likelihood to generate posterior samples, jointly updating latent states and uncertain parameters. This design avoids repeated full-state simulation during forecasting and full-field reconstruction during likelihood evaluation. LD-EnFF substantially outperforms a broad range of data assimilation algorithms on benchmarks spanning Kolmogorov flow, tsunami propagation, and atmospheric modeling, all featuring complex dynamics and sparse, noisy observations.
\end{abstract}

\section{Introduction}
Data assimilation (DA) \citep{sanzalonso2023inverse} combines imperfect model predictions with noisy and incomplete observations to estimate the evolving state of a dynamical system, with applications in weather prediction, oceanography, and geophysical sciences \citep{schneider2022esa,bertino2003sequential,reichle2008data}. Classical approaches include Kalman and ensemble filters \citep{kalman1960new,evensen1994sequential,hunt2007efficient}, particle filters \citep{kunsch2013particle,snyder2008obstacles}, and variational methods \citep{rabier2003variational,desroziers2014link}. Their effectiveness can be limited in high-dimensional nonlinear and non-Gaussian systems by finite-ensemble errors, particle degeneracy, or the cost of repeated model integration and optimization. These challenges are further compounded by sparse observations, which provide only limited direct information about the full state.

Recent generative methods provide flexible alternatives for nonlinear and non-Gaussian filtering. The Ensemble Score Filter (EnSF) estimates the prior score from the predictive ensemble and combines it with the likelihood score to obtain the posterior score \citep{bao2024ensemble,bao2024score}, while the Ensemble Flow Filter (EnFF) constructs flow-matching vector fields from predictive ensembles without training a separate generative model at each assimilation time \citep{lipman2023flow,albergo2023building,liu2023flow,transue2025flow}. However, when applied in the full physical space, both forecast propagation and filtering remain high-dimensional. 

This motivates performing data assimilation in a reduced latent space, particularly for high-dimensional systems with sparse observations. Latent-EnSF carries out the EnSF analysis in a shared latent representation of the state and observations, mitigating the loss of likelihood-score information in unobserved state components, while still propagating the dynamics in the full physical space \citep{si2025latentensf}. LD-EnSF extends this idea by performing both forecast propagation and score-based assimilation in latent space, thereby eliminating full-state dynamical simulation during online assimilation \citep{xiao2026ldensf}. However, it still relies on reverse-time score-based sampling and directly maps sparse observation histories to estimates of the latent state and uncertain model parameters, such as the Reynolds number. Such inverse mappings can be ill-posed under sparse observations, motivating instead the learning of a forward observation map from latent states to physical measurements. These considerations motivate a framework that combines latent dynamics with flow-based filtering and an explicit learned observation model for sparse physical measurements.

\begin{figure*}[!htb]
\centering








\includegraphics[width=\textwidth]{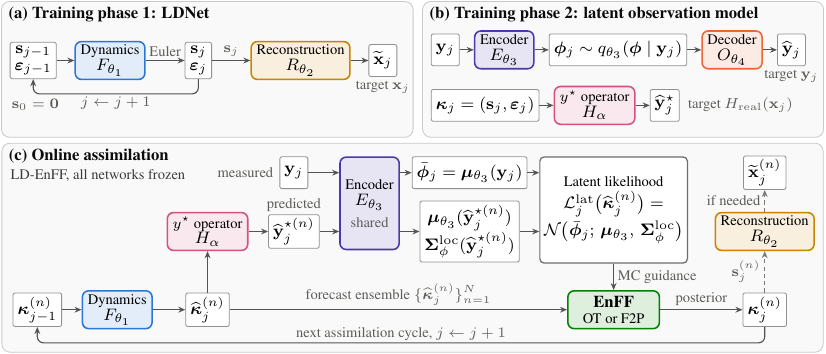}
\caption{
Pipeline of LD-EnFF.
\textbf{Training phase 1:}
LDNet learns parameter-conditioned latent dynamics and a reconstruction map from $\mathbf{s}_j$ to the physical state.
\textbf{Training phase 2:}
A variational encoder maps the sparse noisy observation $\mathbf{y}_j$ to a latent observation representation, while the $y^\star$ operator $H_\alpha$ maps the augmented latent state $\boldsymbol{\kappa}_j=(\mathbf{s}_j,\boldsymbol{\varepsilon}_j)$ to its predicted noiseless observation.
\textbf{Online assimilation:}
With all learned components frozen, LDNet advances the latent states to form the forecast ensemble. EnFF then assimilates each new observation using this ensemble and the state-dependent local Gaussian likelihood in Eq.~\ref{eq:latent-likelihood}, jointly updating the latent states and uncertain parameters. The reconstruction network is invoked only when a physical-state estimate is needed.
}
\label{fig:ld-enff-pipeline}

\end{figure*}

We therefore propose the Latent-Dynamics Ensemble Flow Filter (LD-EnFF), which performs sequential DA in a compact latent space augmented with uncertain parameters. As illustrated in Figure~\ref{fig:ld-enff-pipeline}, we first train an LDNet~\citep{regazzoni2024learning} for latent dynamics and physical-state reconstruction, then learn an observation model linking latent states and uncertain parameters to sparse physical measurements. During online assimilation, LDNet advances the latent ensemble to form a forecast. The learned observation model provides a likelihood that quantifies the agreement of candidate latent states and parameters with each new observation. EnFF then uses this likelihood together with the forecast ensemble to generate posterior samples through a flow-based update, jointly updating the latent states and uncertain parameters. This design avoids full-state numerical simulation during forecasting and repeated full-state reconstruction during likelihood evaluation.

Our main contributions are threefold.
\begin{itemize}
    \item \textbf{Latent-space flow filtering.} We develop LD-EnFF, which couples LDNet-based latent dynamics with EnFF to perform both forecast propagation and analysis in a compact augmented latent space. This enables joint assimilation of the latent dynamical state and uncertain parameter while avoiding repeated full-state simulation during online assimilation.    
    \item \textbf{Learned latent observation modeling.} We combine a variational observation encoder with a forward $y^\star$ operator to construct a state-dependent local Gaussian likelihood that captures both physical observation noise and encoder uncertainty, without requiring repeated full-state reconstruction during EnFF analysis. 
    \item \textbf{Empirical validation.} We evaluate LD-EnFF on Kolmogorov Flow, Tsunami Modeling, and Atmospheric Modeling under sparse and noisy observations. Across all three benchmarks, LD-EnFF consistently achieves the lowest mean final full-state RelRMSE among the methods considered.
\end{itemize}

\section{Background}
\label{sec:background}

This section formulates the data-assimilation problem, reviews EnFF \citep{transue2025flow}, and summarizes related work motivating LD-EnFF.

\subsection{Problem Setup}

We formulate sequential DA as a Bayesian filtering problem \citep{law2015data}. Let $\mathbf{x}_j\in\mathbb{R}^{d_x}$ denote the physical state at assimilation time $j$, and let $\boldsymbol{\varepsilon}\in\mathbb{R}^{d_\varepsilon}$ denote a time-invariant uncertain parameter, such as the Reynolds number in fluid flows. The physical state evolves according to
\begin{equation}
\mathbf{x}_j
=
M(\mathbf{x}_{j-1},\boldsymbol{\varepsilon}),
\qquad
j=1,\ldots,J,
\label{eq:background-physical-dynamics}
\end{equation}
where $M:\mathbb{R}^{d_x}\times\mathbb{R}^{d_\varepsilon}\rightarrow\mathbb{R}^{d_x}$ is a nonlinear forward map. At each assimilation time, observations are generated according to
\begin{equation}
\mathbf{y}_j
=
H_{\mathrm{real}}(\mathbf{x}_j)
+
\boldsymbol{\eta}_j,
\qquad
\boldsymbol{\eta}_j
\overset{\mathrm{i.i.d.}}{\sim}
\mathcal{N}(\mathbf{0},\boldsymbol{\Gamma}_y),
\label{eq:background-observation}
\end{equation}
where $\mathbf{y}_j\in\mathbb{R}^{d_y}$, $H_{\mathrm{real}}:\mathbb{R}^{d_x}\rightarrow\mathbb{R}^{d_y}$ denotes the physical observation operator, with the subscript ``real'' distinguishing it from the learned observation operator $H_{\alpha}$ introduced later in \eqref{eq:ystar-operator}, and $\boldsymbol{\eta}_j$ denotes independent Gaussian observation noise with covariance $\boldsymbol{\Gamma}_y$.

To jointly estimate the physical state and uncertain parameter, we define the augmented state $\boldsymbol{\chi}_j=(\mathbf{x}_j,\boldsymbol{\varepsilon}_j)\in\mathbb{R}^{d_x+d_\varepsilon}$, with $\boldsymbol{\varepsilon}_j=\boldsymbol{\varepsilon}_{j-1}$ during prediction. The corresponding observation likelihood is $p(\mathbf{y}_j\mid\boldsymbol{\chi}_j)=\mathcal{N}(\mathbf{y}_j;H_{\mathrm{real}}(\mathbf{x}_j),\boldsymbol{\Gamma}_y)$.
Given observations $\mathbf{y}_{1:j}=(\mathbf{y}_1,\ldots,\mathbf{y}_j)$, the goal is to infer the joint filtering distribution $p(\boldsymbol{\chi}_j\mid\mathbf{y}_{1:j})=p(\mathbf{x}_j,\boldsymbol{\varepsilon}_j\mid\mathbf{y}_{1:j})$, which quantifies uncertainty in both the physical state and the uncertain parameter. In Bayesian filtering, this distribution is updated recursively through prediction and analysis. The prediction step propagates the previous filtering distribution through the dynamical model to obtain $p(\boldsymbol{\chi}_j\mid\mathbf{y}_{1:j-1})$, while the analysis step incorporates the new observation according to
\begin{equation}
p(\boldsymbol{\chi}_j\mid\mathbf{y}_{1:j})
\propto
p(\mathbf{y}_j\mid\boldsymbol{\chi}_j)
p(\boldsymbol{\chi}_j\mid\mathbf{y}_{1:j-1}).
\label{eq:background-analysis}
\end{equation}
The full prediction and analysis formulas are provided in Appendix~\ref{app:bayesian-filtering}. Although the uncertain parameter $\boldsymbol{\varepsilon}_j=\boldsymbol{\varepsilon}_{j-1}$ remains unchanged during prediction, its posterior distribution is updated during analysis through the observations.

\subsection{Ensemble Flow Filter}
\label{sec:enff}

The Ensemble Flow Filter (EnFF) \citep{transue2025flow} is a training-free sequential Bayesian filtering method based on flow matching. At assimilation time $j$, EnFF first constructs a flow from a chosen reference distribution---for example, a standard Gaussian distribution or the filtering distribution from the previous assimilation time---toward the current predictive distribution $p(\boldsymbol{\chi}_j\mid\mathbf{y}_{1:j-1})$. It then incorporates the new observation $\mathbf{y}_j$ through likelihood-based guidance, modifying the flow so that its terminal distribution targets the filtering posterior $p(\boldsymbol{\chi}_j\mid\mathbf{y}_{1:j})$.

The first step of EnFF is therefore a standard flow-matching construction of the predictive distribution. Let $\boldsymbol{z}_t$, $t\in[0,1]$, denote the flow-time variable. Given sample pairs $\{(\boldsymbol{z}_0^{(n)},\boldsymbol{z}_1^{(n)})\}_{n=1}^N$ drawn from a coupling between the chosen reference distribution and the predictive distribution, the flow-matching marginal vector field is approximated by the Monte Carlo (MC) estimator
\begin{equation}
\boldsymbol{u}_t(\boldsymbol{z})
\approx
\sum_{n=1}^N
w_n(\boldsymbol{z})\,
\boldsymbol{u}_t
\left(
\boldsymbol{z}\mid
\boldsymbol{z}_0^{(n)},
\boldsymbol{z}_1^{(n)}
\right),
\label{eq:enff-marginal-vf}
\end{equation}
where the normalized weights are
$w_n(\boldsymbol{z})
=
\frac{
p_t(\boldsymbol{z}\mid\boldsymbol{z}_0^{(n)},\boldsymbol{z}_1^{(n)})
}{
\sum_{m=1}^N p_t(\boldsymbol{z}\mid\boldsymbol{z}_0^{(m)},\boldsymbol{z}_1^{(m)})
}$.
Here, $p_t(\boldsymbol{z}\mid\boldsymbol{z}_0,\boldsymbol{z}_1)$ denotes a conditional probability path and $\boldsymbol{u}_t(\boldsymbol{z}\mid\boldsymbol{z}_0,\boldsymbol{z}_1)$ its associated conditional vector field. Further details on the flow-matching construction and its MC approximation are provided in Appendix~\ref{app:flow-matching}.

We consider two conditional-flow constructions in our experiments. The optimal-transport (OT) construction uses the Gaussian reference distribution $\mathcal{N}(\mathbf{0},\mathbf{I})$ and transports it toward the predictive distribution along straight conditional paths. In contrast, the filtering-to-predictive (F2P) construction uses the filtering distribution at assimilation time $j-1$ as the reference and directly interpolates toward the predictive distribution at time $j$. For paired endpoints $(\boldsymbol{z}_0,\boldsymbol{z}_1)$, the F2P conditional vector field is
$\boldsymbol{u}_t(\boldsymbol{z}\mid\boldsymbol{z}_0,\boldsymbol{z}_1)=\boldsymbol{z}_1-\boldsymbol{z}_0$.
Thus, F2P directly connects the previous filtering distribution to the current predictive distribution without introducing a fixed Gaussian reference. Following \citet{transue2025flow}, we refer to these two constructions as EnFF-OT and EnFF-F2P, respectively. In LD-EnFF, we use the OT construction directly, whereas for F2P we slightly perturb the source endpoint to enable updates of the time-invariant uncertain parameters during analysis. The explicit conditional probability paths and vector fields are given in Appendix~\ref{app:flow-matching}, while the motivation and construction of the F2P perturbation are detailed in Appendix~\ref{app:f2p-perturbation}.

The second step of EnFF incorporates the current observation through likelihood-based guidance. Specifically, the marginal vector field is augmented as
$\boldsymbol{u}'_t(\boldsymbol{z};\mathbf{y}_j)=\boldsymbol{u}_t(\boldsymbol{z})+\boldsymbol{g}_t(\boldsymbol{z};\mathbf{y}_j)$,
so that the guided flow targets the posterior in \eqref{eq:background-analysis}. The guidance retains the same conditional vector fields used to construct the predictive flow, but reweights the endpoint pairs according to the observation likelihood, thereby shifting the terminal distribution from the predictive distribution to the filtering posterior. EnFF admits both MC and localized guidance approximations, detailed in Appendix~\ref{app:enff-guidance}. Filtering samples are then obtained by integrating the guided vector field from $t=0$ to $t=1$; see Appendix~\ref{app:enff-guidance} for the corresponding flow ODE. The resulting terminal ensemble approximates $p(\boldsymbol{\chi}_j\mid\mathbf{y}_{1:j})$ and is subsequently propagated through the dynamical model to form the predictive ensemble at the next assimilation time.

EnFF avoids training a separate flow model at each assimilation time and supports flexible conditional probability paths, including OT and F2P. However, when applied in a high-dimensional physical state space, both dynamical propagation and filtering remain high-dimensional and can lead to poor data-assimilation performance, as shown in Section~\ref{sec:assimilation-results}. This motivates performing both steps in the learned low-dimensional latent representation to be developed in Section~\ref{sec:method}.

\subsection{Related Work} \label{sec:related-work} 

\noindent\textbf{Generative DA.} Generative models provide flexible tools for nonlinear and non-Gaussian data assimilation. Score-based filtering methods include the score-based filter \citep{bao2024score} and the training-free Ensemble Score Filter (EnSF) \citep{bao2024ensemble}. \citet{rozet2023score} proposed conditional score-based reconstruction of complete state trajectories, formulated as a smoothing problem rather than sequential filtering using only observations available up to the current time. Other recent approaches include FlowDAS \citep{chen2025flowdas}, DiffDA \citep{huang2024diffda}, state-observation augmented diffusion \citep{li2025soad}, and sequential Langevin sampling \citep{ding2026langevin}. While several of these methods require offline training of generative components, EnSF \citep{bao2024ensemble} and EnFF \citep{transue2025flow} construct the score or flow vector field directly from predictive ensembles, avoiding separate generative-model training for the filtering update.

\noindent\textbf{Latent-space DA.} Learned reduced-dimensional representations have been combined with ensemble filtering, surrogate dynamics, and implicit neural representations to reduce the computational cost of high-dimensional DA \citep{peyron2021latent,cheng2023generalised,mucke2024deep,chen2023reduced,li2024latentimplicit}. Latent-EnSF \citep{si2025latentensf} performs EnSF updates in a shared latent representation of states and sparse observations, mitigating the loss of likelihood-score information in unobserved components, while still propagating dynamics in the full state space. LD-EnSF \citep{xiao2026ldensf}, the closest framework to ours, further performs both forecast and filtering in reduced space by combining LDNet-based latent dynamics with EnSF and an LSTM-based encoder for sparse observation histories. However, it still relies on reverse-time score-based sampling and tries to map sparse observation histories to latent-state and uncertain-parameter estimates, an inverse mapping that is generally ill-posed. In contrast, LD-EnFF combines flow-based posterior transport with an explicit forward latent-to-observation model that predicts noiseless sparse physical observations from the augmented latent state. Together with variational observation encoding, this forward model yields a state-dependent latent-space likelihood for assimilating sparse and noisy observations.

\noindent\textbf{Learning-assisted DA.} Deep learning has also been integrated with classical filtering and variational assimilation. Learned surrogate and recurrent models have been used for scalable state estimation \citep{gottwald2021combining,penny2022integrating}, while deep Bayesian filtering learns probabilistic filtering models in latent representations \citep{tarumi2025deep}. Koopman-based approaches incorporate learned latent dynamics into assimilation pipelines \citep{frion2024koopman,singh2024koda}, and learned variational methods such as Tensor-Var improve the scalability of optimization-based assimilation \citep{yang2025tensorvar}. LD-EnFF connects these directions through parameter-conditioned latent dynamics, training-free flow filtering, and an explicit likelihood construction for sparse observations.

\section{Methodology}
\label{sec:method}

This section presents LD-EnFF, comprising LDNet-based latent dynamics, a learned latent observation model, and EnFF analysis in the augmented latent space.

\subsection{Latent Dynamics Networks (LDNets)}
\label{sec:ldnet}

To model the system dynamics in a reduced-dimensional space, we employ a Latent Dynamics Network (LDNet) \citep{regazzoni2024learning}, whose latent dynamics are conditioned on the uncertain parameter $\boldsymbol{\varepsilon}$, as illustrated in Training Phase 1 of Figure~\ref{fig:ld-enff-pipeline}. The model consists of a dynamics network $F_{\theta_1}$ and a reconstruction network $R_{\theta_2}$. Given a latent state $\mathbf{s}_{j-1}\in\mathbb{R}^{d_s}$ and an uncertain parameter $\boldsymbol{\varepsilon}_{j-1}\in\mathbb{R}^{d_\varepsilon}$, the dynamics network predicts the latent-state time derivative as
\begin{equation}
\dot{\mathbf{s}}_{j-1}
=
F_{\theta_1}\!\left(
\mathbf{s}_{j-1},
\boldsymbol{\varepsilon}_{j-1}
\right),
\qquad
j=1,\ldots,J.
\label{eq:ldnet-dynamics}
\end{equation}
The latent state is then advanced using an explicit Euler step,
\begin{equation}
\mathbf{s}_j
=
\mathbf{s}_{j-1}
+
\Delta t\,\dot{\mathbf{s}}_{j-1},
\qquad
\boldsymbol{\varepsilon}_j
=
\boldsymbol{\varepsilon}_{j-1},
\label{eq:ldnet-update}
\end{equation}
where $\Delta t$ denotes the temporal increment between consecutive latent states. The uncertain parameter $\boldsymbol{\varepsilon}$ is assumed to be time-invariant over the assimilation horizon and is therefore carried forward unchanged during dynamical propagation. Different realizations of $\boldsymbol{\varepsilon}$ correspond to different trajectories and therefore induce distinct latent evolutions. The initial latent state is set to $\mathbf{s}_0=\mathbf{0}$.

The reconstruction network maps each latent state back to the corresponding full physical state as $\widetilde{\mathbf{x}}_j=R_{\theta_2}(\mathbf{s}_j)\in\mathbb{R}^{d_x}$.
For a training set of $N_{\mathrm{tr}}$ trajectories, the network parameters $(\theta_1,\theta_2)$ are learned by minimizing the mean squared reconstruction error over all training trajectories
\begin{equation}
\mathcal{L}_{\mathrm{LDNet}}(\theta_1,\theta_2)
=
\frac{1}{N_{\mathrm{tr}}J}
\sum_{i=1}^{N_{\mathrm{tr}}}
\sum_{j=1}^{J}
\left\|
\widetilde{\mathbf{x}}_j^{(i)}
-
\mathbf{x}_j^{(i)}
\right\|_2^2.
\label{eq:ldnet-loss}
\end{equation}
Here, $\mathbf{x}_j^{(i)}$ is the physical state of the $i$th training trajectory at time $j$, generated using its corresponding uncertain-parameter realization $\boldsymbol{\varepsilon}^{(i)}$. The resulting parameter-conditioned latent dynamics provide the reduced-order surrogate for forecast propagation in LD-EnFF.

\subsection{Two-Component Latent Observation Modeling}
\label{sec:latent-observation-model}

To perform the analysis update in a low-dimensional space, we use the augmented latent state $\boldsymbol{\kappa}_j=(\mathbf{s}_j,\boldsymbol{\varepsilon}_j)\in\mathbb{R}^{d_s+d_\varepsilon}$, which combines the latent dynamical state and the uncertain parameter and serves as the filtering variable in LD-EnFF. To construct an observation likelihood for this reduced-dimensional representation, we introduce two learned components, as illustrated in Training Phase 2 of Figure~\ref{fig:ld-enff-pipeline}. A variational observation encoder learns a probabilistic latent representation of the sparse noisy physical observation $\mathbf{y}_j$, while a learned forward $y^\star$ operator maps the augmented latent state $\boldsymbol{\kappa}_j$ to its predicted noiseless sparse physical observation. Section~\ref{sec:local-gaussian} combines these two components to construct the state-dependent latent-observation likelihood used by EnFF.

\subsubsection{Component 1: Variational Observation Encoding}
\label{sec:observation-encoder}

The first component learns a probabilistic latent representation of sparse, noisy physical observations using a variational autoencoder (VAE) \citep{kingma2013auto}. Let $\boldsymbol{\phi}_j\in\mathbb{R}^{d_\phi}$ denote the latent observation variable. The encoder $E_{\theta_3}$ maps $\mathbf{y}_j\in\mathbb{R}^{d_y}$ to a mean vector and an elementwise log-variance vector, $[\boldsymbol{\mu}_{\theta_3}(\mathbf{y}_j),\log\boldsymbol{\sigma}_{\theta_3}^2(\mathbf{y}_j)]=E_{\theta_3}(\mathbf{y}_j)$, which parameterize
\begin{equation}
q_{\theta_3}(\boldsymbol{\phi}_j\mid\mathbf{y}_j)
=
\mathcal{N}\!\left(
\boldsymbol{\phi}_j;
\boldsymbol{\mu}_{\theta_3}(\mathbf{y}_j),
\boldsymbol{\Sigma}_{\mathrm{enc}}(\mathbf{y}_j)
\right),
\qquad
\boldsymbol{\Sigma}_{\mathrm{enc}}(\mathbf{y}_j)
=
\operatorname{diag}\!\left(
\boldsymbol{\sigma}_{\theta_3}^2(\mathbf{y}_j)
\right).
\label{eq:vae-conditional}
\end{equation}
Equivalently, an encoder sample is given by $\boldsymbol{\phi}_j=\boldsymbol{\mu}_{\theta_3}(\mathbf{y}_j)+\boldsymbol{\sigma}_{\theta_3}(\mathbf{y}_j)\odot\boldsymbol{\zeta}$, with $\boldsymbol{\zeta}\sim\mathcal{N}(\mathbf{0},\mathbf{I}_{d_\phi})$. During training, the observation decoder $O_{\theta_4}$ reconstructs the physical observation from latent samples. The VAE training objective and implementation details are provided in Appendix~\ref{app:hyperparameter-choice}.
For filtering, we use the encoder mean $\bar{\boldsymbol{\phi}}_j=\boldsymbol{\mu}_{\theta_3}(\mathbf{y}_j)$ as a deterministic representation of the observation, while retaining the encoder covariance to construct the latent-observation likelihood in Section~\ref{sec:local-gaussian}.

\subsubsection{Component 2: $y^\star$ Operator}
\label{sec:ystar-operator}

The second component learns a forward $y^\star$ operator that maps the augmented latent state directly to the corresponding noiseless sparse physical observation.
This avoids repeatedly reconstructing the full $d_x$-dimensional physical state through $R_{\theta_2}$ and subsequently applying the physical observation operator $H_{\mathrm{real}}$ during likelihood evaluations. To distinguish the noiseless observation from its noisy measurement, we denote $\mathbf{y}_j^\star=H_{\mathrm{real}}(\mathbf{x}_j)$, so that $\mathbf{y}_j=\mathbf{y}_j^\star+\boldsymbol{\eta}_j$ with $\boldsymbol{\eta}_j\sim\mathcal{N}(\mathbf{0},\boldsymbol{\Gamma}_y)$.

We define the learned $y^\star$ operator as
\begin{equation}
H_\alpha:
\mathbb{R}^{d_s+d_\varepsilon}
\rightarrow
\mathbb{R}^{d_y},
\qquad
\widehat{\mathbf{y}}_j^\star
=
H_\alpha(\boldsymbol{\kappa}_j),
\label{eq:ystar-operator}
\end{equation}
which predicts the noiseless sparse physical observation directly from the augmented latent state. Thus, $H_\alpha$ maps the low-dimensional augmented latent space directly to the sparse observation space, bypassing the full physical state space.

Using the augmented latent states obtained from LDNet and the corresponding noiseless observation targets $\mathbf{y}_j^{\star,(i)}=H_{\mathrm{real}}(\mathbf{x}_j^{(i)})$, we learn the operator parameters by minimizing
\begin{equation}
\mathcal{L}_{y^\star}(\alpha)
=
\frac{1}{N_{\mathrm{tr}}J}
\sum_{i=1}^{N_{\mathrm{tr}}}
\sum_{j=1}^{J}
\left\|
H_\alpha(\boldsymbol{\kappa}_j^{(i)})
-
H_{\mathrm{real}}(\mathbf{x}_j^{(i)})
\right\|_2^2.
\label{eq:ystar-loss}
\end{equation}
The predicted noiseless observation is subsequently used to construct the state-dependent latent-observation likelihood in Section~\ref{sec:local-gaussian}.

\subsection{State-Dependent Local Gaussian Latent-Observation Likelihood}
\label{sec:local-gaussian}

For a candidate augmented latent state $\boldsymbol{\kappa}$, the learned $y^\star$ operator predicts the corresponding noiseless physical observation
$\widehat{\mathbf{y}}^\star(\boldsymbol{\kappa})=H_\alpha(\boldsymbol{\kappa})$.
To account for physical observation noise, we consider the corresponding noisy physical-observation variable
$\mathbf{y}=\widehat{\mathbf{y}}^\star(\boldsymbol{\kappa})+\boldsymbol{\eta}$,
where $\boldsymbol{\eta}\sim\mathcal{N}(\mathbf{0},\boldsymbol{\Gamma}_y)$ with density denoted as $q_y$, and propagate it through the trained variational encoder. Marginalizing over the observation noise gives the latent-observation distribution
\begin{equation}
\widetilde{p}(\boldsymbol{\phi}\mid\boldsymbol{\kappa})
:=
\int
q_{\theta_3}\!\left(
\boldsymbol{\phi}
\mid
\widehat{\mathbf{y}}^\star(\boldsymbol{\kappa})+\boldsymbol{\eta}
\right)
q_y(\boldsymbol{\eta}) 
\,d\boldsymbol{\eta}.
\label{eq:latent-mixture-likelihood}
\end{equation}
Since each $q_{\theta_3}(\boldsymbol{\phi}\mid\widehat{\mathbf{y}}^\star(\boldsymbol{\kappa})+\boldsymbol{\eta})$ is Gaussian, \eqref{eq:latent-mixture-likelihood} is a continuous mixture of Gaussians. This mixture would itself be Gaussian if the encoder mean were affine and the encoder covariance constant in the input. For a nonlinear encoder neither holds, so the mixture is generally non-Gaussian and has no closed form. We therefore impose both conditions locally. When the observation noise is small, $\boldsymbol{\eta}$ remains close to zero with high probability, and over this neighborhood of $\widehat{\mathbf{y}}^\star=\widehat{\mathbf{y}}^\star(\boldsymbol{\kappa})$ we linearize the encoder mean and hold the encoder covariance fixed: at
$\widehat{\mathbf{y}}^\star=\widehat{\mathbf{y}}^\star(\boldsymbol{\kappa})$,
\begin{equation}
\boldsymbol{\mu}_{\theta_3}(\widehat{\mathbf{y}}^\star+\boldsymbol{\eta})
\approx
\boldsymbol{\mu}_{\theta_3}(\widehat{\mathbf{y}}^\star)
+
J_\mu(\widehat{\mathbf{y}}^\star)\boldsymbol{\eta},
\qquad
\boldsymbol{\Sigma}_{\mathrm{enc}}(\widehat{\mathbf{y}}^\star+\boldsymbol{\eta})
\approx
\boldsymbol{\Sigma}_{\mathrm{enc}}(\widehat{\mathbf{y}}^\star),
\label{eq:encoder-local-approximation}
\end{equation}
where
$J_\mu(\widehat{\mathbf{y}}^\star)
=
\left.\frac{
\partial\boldsymbol{\mu}_{\theta_3}}{\partial\mathbf{y}}
\right|_{\widehat{\mathbf{y}}^\star}
\in\mathbb{R}^{d_\phi\times d_y}$. Under this approximation, an encoder sample becomes
\begin{equation}
\widetilde{\boldsymbol{\phi}}
=
\boldsymbol{\mu}_{\theta_3}(\widehat{\mathbf{y}}^\star)
+
J_\mu(\widehat{\mathbf{y}}^\star)\boldsymbol{\eta}
+
\boldsymbol{\sigma}_{\theta_3}(\widehat{\mathbf{y}}^\star)
\odot
\boldsymbol{\zeta},
\qquad
\boldsymbol{\zeta}
\sim
\mathcal{N}(\mathbf{0},\mathbf{I}_{d_\phi}),
\quad
\boldsymbol{\zeta}\perp\boldsymbol{\eta}.
\label{eq:local-encoded-sample}
\end{equation}
Since this is an affine transformation of independent Gaussian variables, it induces
\begin{equation}
\widetilde{p}(\boldsymbol{\phi}\mid\boldsymbol{\kappa})
\approx
\mathcal{N}\!\left(
\boldsymbol{\phi};
\boldsymbol{\mu}_{\theta_3}(\widehat{\mathbf{y}}^\star),
\boldsymbol{\Sigma}_{\phi}^{\mathrm{loc}}(\widehat{\mathbf{y}}^\star)
\right),
\label{eq:local-gaussian-surrogate}
\end{equation}
where
\begin{equation}
\boldsymbol{\Sigma}_{\phi}^{\mathrm{loc}}(\widehat{\mathbf{y}}^\star)
=
J_\mu(\widehat{\mathbf{y}}^\star)
\boldsymbol{\Gamma}_y
J_\mu(\widehat{\mathbf{y}}^\star)^\top
+
\boldsymbol{\Sigma}_{\mathrm{enc}}(\widehat{\mathbf{y}}^\star).
\label{eq:local-gaussian-covariance}
\end{equation}
The first term captures physical observation uncertainty propagated through the local encoder Jacobian, while the second term captures encoder uncertainty. Under the same local approximation, the mean and covariance of this Gaussian coincide with the corresponding approximations to the first two moments of the encoder-induced distribution in \eqref{eq:latent-mixture-likelihood}; see Appendix~\ref{app:local-gaussian-moments} for details.

For assimilation, we represent the measured observation in latent-observation space by its encoder mean,
$\bar{\boldsymbol{\phi}}_j
=
\boldsymbol{\mu}_{\theta_3}(\mathbf{y}_j)$. We then define the operational surrogate likelihood by evaluating the local latent-observation density at $\bar{\boldsymbol{\phi}}_j$:
\begin{equation}
\mathcal{L}^{\mathrm{lat}}_j(\boldsymbol{\kappa})
:=
\mathcal{N}\!\left(
\bar{\boldsymbol{\phi}}_j;
\boldsymbol{\mu}_{\theta_3}
\!\left(
\widehat{\mathbf{y}}^\star(\boldsymbol{\kappa})
\right),
\boldsymbol{\Sigma}_{\phi}^{\mathrm{loc}}
\!\left(
\widehat{\mathbf{y}}^\star(\boldsymbol{\kappa})
\right)
\right).
\label{eq:latent-likelihood}
\end{equation}
This likelihood is state-dependent because both its mean and covariance depend on the candidate augmented latent state through
$\widehat{\mathbf{y}}^\star(\boldsymbol{\kappa})
=
H_\alpha(\boldsymbol{\kappa})$.

\subsection{Latent-Dynamics Ensemble Flow Filter (LD-EnFF)}
\label{sec:ld-enff}

After offline training, LD-EnFF alternates between latent forecast propagation and EnFF analysis. At assimilation time $j$, let $\{\boldsymbol{\kappa}_{j-1}^{(n)}\}_{n=1}^{N}$ denote the posterior augmented latent ensemble from the previous cycle, with $\boldsymbol{\kappa}_{j-1}^{(n)}=(\mathbf{s}_{j-1}^{(n)},\boldsymbol{\varepsilon}_{j-1}^{(n)})$. We use hats on $\mathbf{s}_j$, $\boldsymbol{\varepsilon}_j$, and $\boldsymbol{\kappa}_j$ to denote their forecast values before assimilating $\mathbf{y}_j$. Each ensemble member is propagated through the trained LDNet according to
\begin{equation}
\widehat{\mathbf{s}}_j^{(n)}
=
\mathbf{s}_{j-1}^{(n)}
+
\Delta t\,
F_{\theta_1}
\!\left(
\mathbf{s}_{j-1}^{(n)},
\boldsymbol{\varepsilon}_{j-1}^{(n)}
\right),
\qquad
\widehat{\boldsymbol{\varepsilon}}_j^{(n)}
=
\boldsymbol{\varepsilon}_{j-1}^{(n)}.
\label{eq:ld-enff-forecast}
\end{equation}
The resulting forecast augmented latent state is $\widehat{\boldsymbol{\kappa}}_j^{(n)}=(\widehat{\mathbf{s}}_j^{(n)},\widehat{\boldsymbol{\varepsilon}}_j^{(n)})$, and the forecast ensemble provides an empirical approximation to the predictive distribution, whose density is denoted by $\widehat{p}_j(\boldsymbol{\kappa})$.
The incoming observation is encoded as $\bar{\boldsymbol{\phi}}_j=\boldsymbol{\mu}_{\theta_3}(\mathbf{y}_j)$. Using the state-dependent surrogate likelihood $\mathcal{L}^{\mathrm{lat}}_j(\boldsymbol{\kappa})$ from \eqref{eq:latent-likelihood}, the corresponding surrogate analysis density $p_j(\boldsymbol{\kappa})$ satisfies
\begin{equation}
p_j(\boldsymbol{\kappa})
\propto
\mathcal{L}^{\mathrm{lat}}_j(\boldsymbol{\kappa})\,
\widehat{p}_j(\boldsymbol{\kappa}).
\label{eq:ld-enff-analysis}
\end{equation}
EnFF then uses likelihood-based guidance to transport the forecast ensemble toward this analysis distribution, producing the posterior ensemble $\{\boldsymbol{\kappa}_j^{(n)}\}_{n=1}^{N}$ for the next assimilation cycle.

LD-EnFF uses either the OT or F2P conditional-flow construction introduced in Section~\ref{sec:enff}. The OT construction is applied directly. For F2P, however, the exact dynamical pairing gives zero conditional velocity in the parameter coordinates because $\widehat{\boldsymbol{\varepsilon}}_j^{(n)}=\boldsymbol{\varepsilon}_{j-1}^{(n)}$. We therefore slightly perturb the source endpoint before constructing the F2P flow, enabling parameter updates through observation-dependent guidance during analysis. The precise construction and derivation are provided in Appendix~\ref{app:f2p-perturbation}.

Compared with EnSF, the Monte Carlo EnFF guidance used here requires only evaluations of $\mathcal{L}^{\mathrm{lat}}_j$, rather than its score with respect to the augmented latent state. This avoids differentiating the state-dependent likelihood, whose covariance depends on the encoder Jacobian and would therefore introduce higher-order derivatives of the observation encoder.
Moreover, because the likelihood is evaluated entirely through the learned $y^\star$ operator and observation encoder, the (state) reconstruction network $R_{\theta_2}$ is not needed during the EnFF analysis. A physical-space state is reconstructed only when needed as $\widetilde{\mathbf{x}}_j^{(n)}=R_{\theta_2}(\mathbf{s}_j^{(n)})$, for evaluation, visualization, or downstream applications. Thus, repeated likelihood evaluations avoid reconstruction of the full $d_x$-dimensional physical state. A pseudocode summary of one LD-EnFF assimilation step is provided in Appendix~\ref{app:ld-enff-algorithm}.

\section{Numerical Experiments}
\label{sec:experiments}

We evaluate LD-EnFF on three nonlinear DA benchmarks with sparse and noisy observations: Kolmogorov Flow, Tsunami Modeling, and Atmospheric Modeling based on PlanetSWE. Detailed governing equations, data-generation procedures, and simulation settings are provided in Appendix~\ref{app:benchmark-details}.

\noindent\textbf{Kolmogorov Flow.}
We consider turbulent Kolmogorov flow on a $150\times150$ spatial grid, with the Reynolds number treated as the uncertain parameter \citep{kochkov2021machine}. Both velocity components are observed on a fixed regular $10\times10$ grid, corresponding to $0.44\%$ of the spatial locations, at 40 assimilation times.
\textbf{Tsunami Modeling.}
We consider two-dimensional shallow-water wave propagation following the setup of \citet{si2025latentensf}, with uncertainty in the spatial location of the initial Gaussian surface perturbation. The system is discretized on a $150\times150$ grid, and all three state components are observed on a fixed regular $10\times10$ grid ($0.44\%$ of the spatial locations) at 50 assimilation times.
\textbf{Atmospheric Modeling.}
We consider PlanetSWE \citep{ohana2024well}, a rotating shallow-water model on the sphere with Earth-like topography and uncertain forcing parameters. The physical state has dimension $512\times256\times3=393{,}216$, and all three components are observed on a regular $16\times8$ grid (approximately $0.1\%$ of the spatial locations) at 63 assimilation times.

\subsection{Compared Methods and Evaluation Protocol}
\label{sec:experimental-protocol}

For the proposed LD-EnFF, we consider the OT and F2P conditional-flow constructions in Section~\ref{sec:enff}, denoted LD-EnFF-OT and LD-EnFF-F2P, respectively. We also evaluate LD-IEnKF, which retains the proposed LDNet dynamics and latent-observation model but replaces EnFF with IEnKF \citep{sakov2012iterative}, to assess the framework with an alternative (classical Kalman-type) analysis method. 

To isolate the contribution of the latent-observation construction, we compare against LD-EnSF \citep{xiao2026ldensf}, denoted LDNet+LSTM EnSF, which uses the same LDNet dynamics but an LSTM-based observation model; we further construct matched LDNet+LSTM EnFF-OT/F2P baselines that retain the same LDNet dynamics and EnFF analysis as (our) LD-EnFF-OT/F2P while replacing only the latent-observation model.

Additional baselines include LDNet dynamics only as a no-assimilation reference, LETKF \citep{hunt2007efficient} as a full-space baseline, and full-space EnFF-OT/F2P for Kolmogorov flow. Full-space EnFF is not reported for tsunami and atmospheric modeling because the corresponding runs exhibited severe error growth under sparse observations. A summary of the compared methods and their main components is provided in Appendix~\ref{app:compared-methods}.

Within each benchmark, all methods are evaluated on the same randomly selected set of 20 held-out test trajectories, with three random seeds used for each trajectory. For each trajectory and seed, all compared methods use the same sparse noisy observations. All ensemble-based methods use $N=20$ ensemble members, and the true uncertain parameter associated with each test trajectory is not used for initialization. The same Gaussian observation-noise covariance, $\boldsymbol{\Gamma}_y=0.1^2\mathbf{I}_{d_y}$, is used across all three benchmarks. All reported LD-EnFF experiments use the Monte Carlo EnFF guidance described in Appendix~\ref{app:enff-guidance}. Further training and initialization details are provided in Appendix~\ref{app:evaluation-protocol}.
For latent-space methods, the physical-state estimate is obtained by averaging the reconstructed posterior ensemble,
$\widehat{\mathbf{x}}_j
=
N^{-1}\sum_{n=1}^{N}
R_{\theta_2}(\mathbf{s}_j^{(n)})$;
for full-space methods, $\widehat{\mathbf{x}}_j$ is the corresponding ensemble mean. We evaluate assimilation accuracy using the full-state relative root-mean-square error (RelRMSE), calculated as  $\operatorname{RelRMSE}_j(\%)=100\|\widehat{\mathbf{x}}_j-\mathbf{x}_j\|_2/\|\mathbf{x}_j\|_2$.

\subsection{Comparison of Assimilation Accuracy}
\label{sec:assimilation-results}

\begin{figure*}[t]
\centering
\includegraphics[width=\textwidth]{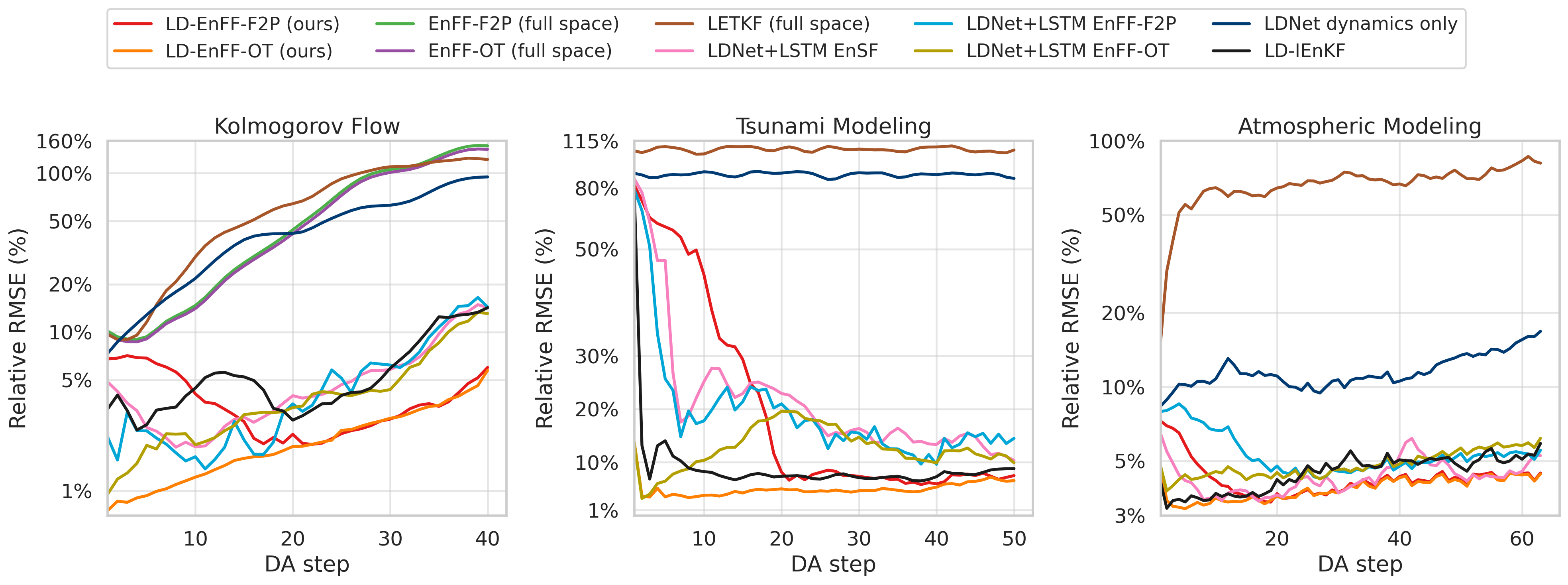}
\caption{Mean full-state RelRMSE over 20 held-out test trajectories and three random seeds per trajectory for Kolmogorov Flow (left), Tsunami Modeling (center), and Atmospheric Modeling (right). Full-space EnFF methods are not shown for Tsunami Modeling and Atmospheric Modeling because the corresponding runs exhibited severe error growth under the sparse-observation setting.}
\label{fig:all-datasets-results}
\end{figure*}

Figure~\ref{fig:all-datasets-results} shows the mean full-state RelRMSE over the assimilation horizon for the three benchmarks, while Table~\ref{tab:latent-final-rmse} reports the corresponding full-state RelRMSE at the final assimilation step for the latent-space methods as mean $\pm$ standard deviation.

\begin{table*}[!htb]
\centering
\caption{Full-state RelRMSE (\%) at the final assimilation step for the latent-space methods, reported as mean $\pm$ standard deviation over 20 held-out test trajectories after averaging over three random seeds per trajectory. Best and second-best results are shown in bold and underlined, respectively.}
\label{tab:latent-final-rmse}
\small
\begin{tabular}{lccc}
\toprule
Method & Kolmogorov Flow & Tsunami Modeling & Atmospheric Modeling \\
\midrule
LD-EnFF-OT (ours)
& $\bm{5.74 \pm 0.26}$
& $\bm{6.53 \pm 0.25}$
& $\bm{4.42 \pm 0.21}$ \\

LD-EnFF-F2P (ours)
& \underline{$5.99 \pm 0.29$}
& \underline{$7.48 \pm 0.26$}
& \underline{$4.46 \pm 0.23$} \\

LD-IEnKF
& $14.23 \pm 0.47$
& $8.79 \pm 0.45$
& $5.89 \pm 0.42$ \\

LDNet+LSTM EnSF
& $14.27 \pm 0.49$
& $10.35 \pm 0.42$
& $5.26 \pm 0.31$ \\

LDNet+LSTM EnFF-OT
& $13.09 \pm 0.38$
& $9.87 \pm 0.32$
& $6.18 \pm 0.27$ \\

LDNet+LSTM EnFF-F2P
& $14.40 \pm 0.42$
& $14.48 \pm 0.39$
& $5.52 \pm 0.29$ \\
\bottomrule
\end{tabular}
\end{table*}

As summarized in Table~\ref{tab:latent-final-rmse}, LD-EnFF-OT and LD-EnFF-F2P achieve the lowest and second-lowest mean final full-state RelRMSE across all three benchmarks. Figure~\ref{fig:all-datasets-results} further shows faster early-stage error reduction for LD-EnFF-OT. The uncertain-parameter trajectories in Appendix~\ref{app:visualization} exhibit the same trend, with OT reducing the parameter error more rapidly at early stages. This is consistent with the conditional-flow constructions: exact dynamical F2P has zero velocity in the parameter coordinates because the forecast leaves $\boldsymbol{\varepsilon}$ unchanged, whereas the perturbed F2P coupling introduces $O(\sigma_{\min})$ parameter-space directions that can be reweighted by the observation-dependent MC guidance. OT is not subject to this constraint. More broadly, all assimilating latent-space methods substantially reduce the uncertain-parameter error, supporting joint updates of the latent state and $\boldsymbol{\varepsilon}$.

The matched LDNet+LSTM EnFF-OT and LDNet+LSTM EnFF-F2P baselines yield higher mean final errors than their LD-EnFF counterparts, supporting the proposed latent-observation construction when the LDNet dynamics and EnFF analysis are fixed. LD-IEnKF, which retains the same LDNet dynamics and latent-observation model but replaces EnFF with IEnKF, still substantially improves over the no-assimilation LDNet reference, showing that the learned latent formulation remains effective with an alternative analysis method, although its final errors remain higher than those of LD-EnFF. LDNet+LSTM EnSF also yields higher errors than LD-EnFF, while the much larger errors of the LDNet dynamics-only reference further highlight the importance of observational correction. Figure~\ref{fig:all-datasets-results} also shows that the reported full-space baselines are substantially less accurate under sparse observations. Final-state visualizations are provided in Appendix~\ref{app:visualization}.

\section{Conclusion and Future Work}
\label{sec:conclusion}

We introduced LD-EnFF, a latent-space framework for high-dimensional nonlinear data assimilation with sparse and noisy observations. It combines LDNet-based latent dynamics, Ensemble Flow Filtering, and a learned latent observation model to perform forecast propagation and analysis in an augmented latent space while jointly updating the latent state and uncertain parameter. The resulting state-dependent local Gaussian likelihood incorporates observation noise and encoder uncertainty without repeated full-state reconstruction. Across all three benchmarks, LD-EnFF achieves the lowest mean final full-state RelRMSE among the compared latent-space methods.
The framework is modular, allowing LDNet, the $y^\star$ operator, and the variational observation encoder to be replaced by alternative components. Future work will explore end-to-end assimilation-aware training, alternative EnFF guidance approximations, and more complex, higher-dimensional parameter spaces.

\subsection*{AI use statement}

Generative AI tools were used to assist with drafting portions of the manuscript and editing the text for clarity and readability. They were not used for research methodology or experimental design, formulation or proof of mathematical claims, implementation of methods, generation or processing of datasets, or interpretation of experimental results. All AI-assisted text was reviewed and revised by the authors. The authors take full responsibility for the final content of this work.

\subsection*{Reproducibility statement}

Benchmark definitions and data-generation procedures are provided in Appendix~\ref{app:benchmark-details}. Model architectures, normalization procedures, training hyperparameters, and implementation setup are documented in Appendix~\ref{app:training-details}. The experimental protocol, including data splits, observation generation, ensemble initialization, and evaluation settings, is described in Section~\ref{sec:experimental-protocol} and Appendix~\ref{app:evaluation-protocol}. Additional derivations and algorithmic details for the filtering procedure are provided in Appendices~\ref{app:additional-background} and~\ref{app:method-details}.

\subsection*{Acknowledgments}

Z.T., K.S., W.H., and W.Z. acknowledge support from the National Science Foundation under Awards DMS-2502900 and DMS-2540370 and from the Air Force Office of Scientific Research under Grant No. FA9550-25-1-0079. P.S. and P.C. acknowledge support from the National Science Foundation under Awards DMS-2245111 and CNS-2325631 and from the U.S. Department of Energy under Grant Nos. DE-AC05-00OR22725 and DE-AC02-06CH11357.

\bibliography{iclr2027_conference}
\bibliographystyle{iclr2027_conference}

\clearpage
\appendix

\section*{Appendix}

\section{Additional Background}
\label{app:additional-background}

\subsection{Bayesian Filtering Recursion}
\label{app:bayesian-filtering}

For $j=1,\ldots,J$, given the filtering distribution at the previous assimilation time, the predictive distribution is
\begin{equation}
p(\boldsymbol{\chi}_j\mid\mathbf{y}_{1:j-1})
=
\int
p(\boldsymbol{\chi}_j\mid\boldsymbol{\chi}_{j-1})
p(\boldsymbol{\chi}_{j-1}\mid\mathbf{y}_{1:j-1})
\,d\boldsymbol{\chi}_{j-1},
\label{eq:background-prediction}
\end{equation}
where $p(\boldsymbol{\chi}_j\mid\boldsymbol{\chi}_{j-1})$ is the transition law induced by the physical dynamics in \eqref{eq:background-physical-dynamics}, with the parameter component carried forward unchanged as $\boldsymbol{\varepsilon}_j=\boldsymbol{\varepsilon}_{j-1}$.

After receiving the observation $\mathbf{y}_j$, the filtering distribution is updated by Bayes' rule,
\begin{equation}
p(\boldsymbol{\chi}_j\mid\mathbf{y}_{1:j})
=
\frac{1}{Z_j}
p(\mathbf{y}_j\mid\boldsymbol{\chi}_j)
p(\boldsymbol{\chi}_j\mid\mathbf{y}_{1:j-1}),
\label{eq:appendix-background-analysis}
\end{equation}
where
\begin{equation}
Z_j
=
\int
p(\mathbf{y}_j\mid\boldsymbol{\chi}_j)
p(\boldsymbol{\chi}_j\mid\mathbf{y}_{1:j-1})
\,d\boldsymbol{\chi}_j
\end{equation}
is the normalization constant.

\subsection{Flow Matching and EnFF Conditional Flows}
\label{app:flow-matching}

Flow matching constructs a time-dependent vector field that transports samples from a reference distribution $\rho_0$ toward a target distribution $\rho_1$. In EnFF \citep{transue2025flow}, the target distribution $\rho_1$ is the predictive distribution at the current assimilation time. Rather than learning this marginal vector field with a separate neural network at every assimilation time, EnFF constructs it directly from conditional flows associated with pairs of reference and target samples.

Let $p(\boldsymbol{z}_0,\boldsymbol{z}_1)$ denote a coupling with marginals $\rho_0$ and $\rho_1$. For each endpoint pair $(\boldsymbol{z}_0,\boldsymbol{z}_1)$, consider a conditional probability path $p_t(\boldsymbol{z}\mid\boldsymbol{z}_0,\boldsymbol{z}_1)$ and an associated conditional vector field $\boldsymbol{u}_t(\boldsymbol{z}\mid\boldsymbol{z}_0,\boldsymbol{z}_1)$ that generates this path. The induced marginal probability path is
\begin{equation}
p_t(\boldsymbol{z})
=
\iint
p_t(\boldsymbol{z}\mid\boldsymbol{z}_0,\boldsymbol{z}_1)
p(\boldsymbol{z}_0,\boldsymbol{z}_1)
\,d\boldsymbol{z}_0\,d\boldsymbol{z}_1.
\label{eq:appendix-flow-marginal-path}
\end{equation}
The corresponding marginal vector field is
\begin{equation}
\boldsymbol{u}_t(\boldsymbol{z})
=
\iint
\boldsymbol{u}_t(\boldsymbol{z}\mid\boldsymbol{z}_0,\boldsymbol{z}_1)
\frac{
p_t(\boldsymbol{z}\mid\boldsymbol{z}_0,\boldsymbol{z}_1)
p(\boldsymbol{z}_0,\boldsymbol{z}_1)
}{
p_t(\boldsymbol{z})
}
\,d\boldsymbol{z}_0\,d\boldsymbol{z}_1,
\label{eq:appendix-flow-marginal-vf}
\end{equation}
or equivalently,
\begin{equation}
\boldsymbol{u}_t(\boldsymbol{z})
=
\mathbb{E}\!\left[
\boldsymbol{u}_t(\boldsymbol{z}\mid\boldsymbol{z}_0,\boldsymbol{z}_1)
\,\middle|\,
\boldsymbol{z}_t=\boldsymbol{z}
\right].
\label{eq:appendix-flow-conditional-expectation}
\end{equation}
By construction, $\boldsymbol{u}_t$ generates the marginal probability path $p_t$ that transports the reference distribution toward the target distribution. Given endpoint pairs $\{(\boldsymbol{z}_0^{(n)},\boldsymbol{z}_1^{(n)})\}_{n=1}^N$ sampled from the coupling, EnFF approximates this marginal vector field using the Monte Carlo estimator in \eqref{eq:enff-marginal-vf}, avoiding the need to train a separate flow model at every assimilation time. The specific EnFF construction is determined by the choice of reference distribution, endpoint coupling, and conditional probability path. We next summarize the OT and F2P constructions used in this work.

\noindent\textbf{Optimal-transport conditional flow.}
For the OT construction, the reference distribution is $\rho_0=\mathcal{N}(\mathbf{0},\mathbf{I})$, and the conditional probability path is
\begin{equation}
p_t(\boldsymbol{z}_t\mid\boldsymbol{z}_1)
=
\mathcal{N}\!\left(
\boldsymbol{z}_t;
t\boldsymbol{z}_1,
[1-(1-\sigma_{\min})t]^2\mathbf{I}
\right),
\qquad
\sigma_{\min}>0.
\label{eq:enff-ot}
\end{equation}
The corresponding conditional vector field is
\begin{equation}
\boldsymbol{u}_t(\boldsymbol{z}\mid\boldsymbol{z}_1)
=
\frac{
\boldsymbol{z}_1-(1-\sigma_{\min})\boldsymbol{z}
}{
1-(1-\sigma_{\min})t
}.
\label{eq:appendix-enff-ot-vf}
\end{equation}

\noindent\textbf{Filtering-to-predictive conditional flow.}
For F2P, the filtering distribution at assimilation time $j-1$ is used as the reference distribution, while the predictive distribution at time $j$ is the target distribution. Given paired endpoints $(\boldsymbol{z}_0,\boldsymbol{z}_1)$, the conditional probability path is
\begin{equation}
p_t(\boldsymbol{z}_t\mid\boldsymbol{z}_0,\boldsymbol{z}_1)
=
\mathcal{N}\!\left(
\boldsymbol{z}_t;
(1-t)\boldsymbol{z}_0+t\boldsymbol{z}_1,
\sigma_{\min}^2\mathbf{I}
\right),
\qquad
\sigma_{\min}>0.
\label{eq:enff-f2p}
\end{equation}
The corresponding conditional vector field is
\begin{equation}
\boldsymbol{u}_t(\boldsymbol{z}\mid\boldsymbol{z}_0,\boldsymbol{z}_1)
=
\boldsymbol{z}_1-\boldsymbol{z}_0.
\label{eq:appendix-enff-f2p-vf}
\end{equation}

In all reported experiments, we set $\sigma_{\min}=0.001$ for both the OT and F2P conditional flows.

The OT and F2P constructions above specify the conditional probability paths and vector fields used to construct the marginal vector field $\boldsymbol{u}_t$. To incorporate the current observation and obtain the filtering posterior, EnFF further modifies this marginal vector field through likelihood-based guidance, as described in Appendix~\ref{app:enff-guidance}.

\subsection{EnFF Guidance}
\label{app:enff-guidance}

We describe the likelihood-based guidance used by EnFF \citep{transue2025flow}, together with the coupling-dependent formulation introduced by \citet{feng2025on}. Suppose the observation likelihood associated with the flow endpoint $\boldsymbol{z}_1$ can be written as
\begin{equation}
p(\mathbf{y}_j\mid\boldsymbol{z}_1)
\propto
\exp\!\left[-J(\boldsymbol{z}_1;\mathbf{y}_j)\right],
\label{eq:appendix-enff-likelihood}
\end{equation}
where $J(\boldsymbol{z}_1;\mathbf{y}_j)$ denotes the negative log-likelihood up to an additive constant. If $\rho_1$ denotes the predictive endpoint density, the corresponding filtering endpoint density is therefore proportional to $\rho_1(\boldsymbol{z}_1)\exp[-J(\boldsymbol{z}_1;\mathbf{y}_j)]$.

Let $\pi(\boldsymbol{z}_0,\boldsymbol{z}_1)$ denote the original endpoint coupling, and let $\pi'(\boldsymbol{z}_0,\boldsymbol{z}_1)$ denote the endpoint coupling for the guided flow, with the same reference marginal and the filtering endpoint marginal defined above. The reverse-coupling ratio is
\begin{equation}
\mathcal{P}(\boldsymbol{z}_0,\boldsymbol{z}_1)
=
\frac{
\pi'(\boldsymbol{z}_0\mid\boldsymbol{z}_1)
}{
\pi(\boldsymbol{z}_0\mid\boldsymbol{z}_1)
}.
\label{eq:appendix-reverse-coupling-ratio}
\end{equation}
Retaining the same conditional probability paths and conditional vector fields, the guidance term is
\begin{equation}
\boldsymbol{g}_t(\boldsymbol{z}_t;\mathbf{y}_j)
=
\mathbb{E}\!\left[
\left(
\frac{
\mathcal{P}(\boldsymbol{z}_0,\boldsymbol{z}_1)
e^{-J(\boldsymbol{z}_1;\mathbf{y}_j)}
}{
Z_t(\boldsymbol{z}_t;\mathbf{y}_j)
}
-1
\right)
\boldsymbol{u}_t(\boldsymbol{z}_t\mid\boldsymbol{z}_0,\boldsymbol{z}_1)
\,\middle|\,
\boldsymbol{z}_t
\right],
\label{eq:appendix-enff-guidance}
\end{equation}
where
\begin{equation}
Z_t(\boldsymbol{z}_t;\mathbf{y}_j)
=
\mathbb{E}\!\left[
\mathcal{P}(\boldsymbol{z}_0,\boldsymbol{z}_1)
e^{-J(\boldsymbol{z}_1;\mathbf{y}_j)}
\,\middle|\,
\boldsymbol{z}_t
\right].
\label{eq:appendix-enff-normalizer}
\end{equation}
The resulting guided vector field is
$\boldsymbol{u}'_t(\boldsymbol{z}_t;\mathbf{y}_j)
=
\boldsymbol{u}_t(\boldsymbol{z}_t)
+
\boldsymbol{g}_t(\boldsymbol{z}_t;\mathbf{y}_j)$.

For an independent endpoint coupling,
$\pi(\boldsymbol{z}_0,\boldsymbol{z}_1)
=
\rho_0(\boldsymbol{z}_0)\rho_1(\boldsymbol{z}_1)$,
the reverse-coupling ratio satisfies $\mathcal{P}=1$ exactly. For dependent couplings, $\mathcal{P}=1$ is generally not exact, although $\mathcal{P}\approx1$ can be a reasonable approximation for some practical dependent couplings \citep{feng2025on}. In this work, we use the approximation $\mathcal{P}\approx1$.

For F2P, the coupling choice introduces a practical tradeoff. An independent endpoint coupling makes $\mathcal{P}=1$ exact, but removes the memberwise association between each previous filtering member and its own forecast; this variant performed poorly in our experiments. In contrast, the standard dynamical F2P coupling retains this memberwise association, but because the uncertain parameter is unchanged during prediction, it produces zero conditional velocity in the parameter coordinates. We therefore use the perturbed F2P coupling described in Appendix~\ref{app:f2p-perturbation}, which retains the memberwise pairing structure while slightly randomizing the source endpoint to avoid this degeneracy. Since the perturbed F2P coupling remains dependent, $\mathcal{P}=1$ is treated as a practical approximation rather than an exact identity.

Since the guidance term in \eqref{eq:appendix-enff-guidance} is generally intractable, EnFF uses a Monte Carlo estimator and a localized approximation. The expressions below use the approximation $\mathcal{P}\approx1$ adopted in this work.

\noindent\textbf{Monte Carlo guidance.}
Given sample pairs $\{(\boldsymbol{z}_0^{(n)},\boldsymbol{z}_1^{(n)})\}_{n=1}^N$, the guided vector field is approximated by
\begin{equation}
\boldsymbol{u}'_t(\boldsymbol{z};\mathbf{y}_j)
\approx
\sum_{n=1}^N
w_n'(\boldsymbol{z};\mathbf{y}_j)
\boldsymbol{u}_t
\left(
\boldsymbol{z}\mid
\boldsymbol{z}_0^{(n)},
\boldsymbol{z}_1^{(n)}
\right),
\label{eq:appendix-enff-mc-guidance}
\end{equation}
where
\begin{equation}
w_n'(\boldsymbol{z};\mathbf{y}_j)
=
\frac{
e^{-J(\boldsymbol{z}_1^{(n)};\mathbf{y}_j)}
p_t(\boldsymbol{z}\mid\boldsymbol{z}_0^{(n)},\boldsymbol{z}_1^{(n)})
}{
\sum_{m=1}^N
e^{-J(\boldsymbol{z}_1^{(m)};\mathbf{y}_j)}
p_t(\boldsymbol{z}\mid\boldsymbol{z}_0^{(m)},\boldsymbol{z}_1^{(m)})
}.
\label{eq:appendix-enff-mc-weights}
\end{equation}

\noindent\textbf{Localized guidance.}
Let
$\widehat{\boldsymbol{z}}_1
=
\mathbb{E}[\boldsymbol{z}_1\mid\boldsymbol{z}_t]$
denote the conditional prediction of the flow endpoint. Under the approximation $\mathcal{P}\approx1$, linearizing the negative log-likelihood around $\widehat{\boldsymbol{z}}_1$ yields
\begin{equation}
\boldsymbol{g}_t(\boldsymbol{z}_t;\mathbf{y}_j)
\approx
-
\mathbb{E}\!\left[
\boldsymbol{u}_t(\boldsymbol{z}_t\mid\boldsymbol{z}_0,\boldsymbol{z}_1)
(\boldsymbol{z}_1-\widehat{\boldsymbol{z}}_1)^\top
\,\middle|\,
\boldsymbol{z}_t
\right]
\nabla_{\widehat{\boldsymbol{z}}_1}
J(\widehat{\boldsymbol{z}}_1;\mathbf{y}_j).
\label{eq:appendix-enff-localized-guidance}
\end{equation}

For affine conditional flows of the form
\begin{equation}
\boldsymbol{z}_t
=
\alpha_t\boldsymbol{z}_1
+
\beta_t\boldsymbol{z}_0
+
\sigma_t\boldsymbol{\omega},
\qquad
\boldsymbol{\omega}\sim\mathcal{N}(\mathbf{0},\mathbf{I}),
\label{eq:appendix-enff-affine-flow}
\end{equation}
the localized guidance can be written as
\begin{equation}
\boldsymbol{g}_t(\boldsymbol{z}_t;\mathbf{y}_j)
\approx
-
\boldsymbol{\Lambda}_t
\nabla_{\widehat{\boldsymbol{z}}_1}
J(\widehat{\boldsymbol{z}}_1;\mathbf{y}_j),
\label{eq:appendix-enff-localized-affine}
\end{equation}
where
\begin{equation}
\boldsymbol{\Lambda}_t
=
\frac{
\dot{\alpha}_t\beta_t-\dot{\beta}_t\alpha_t
}{
\beta_t
}
\boldsymbol{\Sigma}_{1\mid t},
\qquad
\boldsymbol{\Sigma}_{1\mid t}
=
\mathbb{E}\!\left[
(\boldsymbol{z}_1-\widehat{\boldsymbol{z}}_1)
(\boldsymbol{z}_1-\widehat{\boldsymbol{z}}_1)^\top
\mid
\boldsymbol{z}_t
\right].
\label{eq:appendix-enff-localized-covariance}
\end{equation}
Both the OT and F2P conditional flows described in Section~\ref{sec:enff} satisfy this affine-flow form. Since evaluating $\boldsymbol{\Sigma}_{1\mid t}$ is generally intractable, EnFF uses the practical approximation
$\boldsymbol{\Lambda}_t\approx\lambda\mathbf{I}$ with $\lambda>0$, giving
\begin{equation}
\boldsymbol{g}_t(\boldsymbol{z}_t;\mathbf{y}_j)
\approx
-\lambda
\nabla_{\widehat{\boldsymbol{z}}_1}
J(\widehat{\boldsymbol{z}}_1;\mathbf{y}_j).
\label{eq:appendix-enff-localized-practical}
\end{equation}

\noindent\textbf{Guidance choice in LD-EnFF.}
For LD-EnFF, we use Monte Carlo guidance in all reported experiments. Localized guidance would require differentiating the negative log-likelihood of the proposed latent-space likelihood with respect to the augmented latent state. In particular, the state-dependent covariance
$\boldsymbol{\Sigma}_{\phi}^{\mathrm{loc}}$
in \eqref{eq:local-gaussian-covariance} depends on the encoder Jacobian $J_\mu$, so differentiating the likelihood introduces higher-order derivatives through the observation encoder. In contrast, Monte Carlo guidance requires only evaluations of the latent-space likelihood
$\mathcal{L}^{\mathrm{lat}}_j(\boldsymbol{\kappa})$
defined in \eqref{eq:latent-likelihood}. We therefore use Monte Carlo guidance throughout the LD-EnFF experiments.

Once the guided vector field $\boldsymbol{u}'_t$ is constructed, filtering samples are obtained by solving
\begin{equation}
\frac{d\boldsymbol{z}_t}{dt}
=
\boldsymbol{u}'_t(\boldsymbol{z}_t;\mathbf{y}_j),
\qquad
t\in[0,1].
\label{eq:enff-flow-ode}
\end{equation}
Starting from samples drawn from the reference distribution at $t=0$, we integrate this ODE to $t=1$ using the explicit Euler method. The resulting terminal ensemble approximates the filtering posterior.

\section{Additional Methodological Details}
\label{app:method-details}

\subsection{Moment Verification for the Local Gaussian Approximation}
\label{app:local-gaussian-moments}

We verify that the mean and covariance of the locally induced Gaussian in \eqref{eq:local-gaussian-surrogate} coincide with the corresponding approximations to the first two moments of the encoder-induced distribution in \eqref{eq:latent-mixture-likelihood} under the local assumptions in \eqref{eq:encoder-local-approximation}. Let
$\widehat{\mathbf{y}}^\star=\widehat{\mathbf{y}}^\star(\boldsymbol{\kappa})$
and
$\boldsymbol{\eta}\sim\mathcal{N}(\mathbf{0},\boldsymbol{\Gamma}_y)$.
By the laws of total expectation and total covariance,
\begin{equation}
\begin{aligned}
\mathbb{E}_{\widetilde{p}(\boldsymbol{\phi}\mid\boldsymbol{\kappa})}
[\boldsymbol{\phi}]
&=
\mathbb{E}_{\boldsymbol{\eta}}
\left[
\boldsymbol{\mu}_{\theta_3}
\left(
\widehat{\mathbf{y}}^\star+\boldsymbol{\eta}
\right)
\right]
\approx
\boldsymbol{\mu}_{\theta_3}
\left(
\widehat{\mathbf{y}}^\star
\right),
\\
\operatorname{Cov}_{\widetilde{p}(\boldsymbol{\phi}\mid\boldsymbol{\kappa})}
(\boldsymbol{\phi})
&=
\mathbb{E}_{\boldsymbol{\eta}}
\left[
\boldsymbol{\Sigma}_{\mathrm{enc}}
\left(
\widehat{\mathbf{y}}^\star+\boldsymbol{\eta}
\right)
\right]
+
\operatorname{Cov}_{\boldsymbol{\eta}}
\left(
\boldsymbol{\mu}_{\theta_3}
\left(
\widehat{\mathbf{y}}^\star+\boldsymbol{\eta}
\right)
\right)
\\
&\approx
\boldsymbol{\Sigma}_{\mathrm{enc}}
\left(
\widehat{\mathbf{y}}^\star
\right)
+
J_\mu
\left(
\widehat{\mathbf{y}}^\star
\right)
\boldsymbol{\Gamma}_y
J_\mu
\left(
\widehat{\mathbf{y}}^\star
\right)^\top
\\
&=
\boldsymbol{\Sigma}_{\phi}^{\mathrm{loc}}
\left(
\widehat{\mathbf{y}}^\star
\right).
\end{aligned}
\label{eq:appendix-local-moments}
\end{equation}

The mean approximation follows from the local linearization in \eqref{eq:encoder-local-approximation} and
$\mathbb{E}[\boldsymbol{\eta}]=\mathbf{0}$.
The covariance approximation follows from the same linearization and the local covariance-freezing assumption, together with
$\operatorname{Cov}(\boldsymbol{\eta})=\boldsymbol{\Gamma}_y$.
Thus, the Gaussian approximation in \eqref{eq:local-gaussian-surrogate} is consistent with the corresponding local approximations to the first two moments of the noise-marginalized encoder-induced distribution in \eqref{eq:latent-mixture-likelihood}.

\subsection{Perturbed F2P Coupling in LD-EnFF}
\label{app:f2p-perturbation}

For the augmented latent state $\boldsymbol{z}=(\mathbf{s},\boldsymbol{\varepsilon})$, the uncertain parameter is unchanged during forecast propagation. Hence, under the exact dynamical F2P coupling at assimilation time $j$,
\begin{equation}
\boldsymbol{z}_0^{(n)}
=
\left(
\mathbf{s}_{j-1}^{(n)},
\boldsymbol{\varepsilon}_{j-1}^{(n)}
\right),
\qquad
\boldsymbol{z}_1^{(n)}
=
\left(
\widehat{\mathbf{s}}_j^{(n)},
\boldsymbol{\varepsilon}_{j-1}^{(n)}
\right).
\label{eq:appendix-f2p-exact-pair}
\end{equation}
Since the standard F2P conditional vector field is $\boldsymbol{u}_t(\boldsymbol{z}\mid\boldsymbol{z}_0,\boldsymbol{z}_1)=\boldsymbol{z}_1-\boldsymbol{z}_0$, its parameter component satisfies
\begin{equation}
\left[
\boldsymbol{u}_t
\left(
\boldsymbol{z}\mid
\boldsymbol{z}_0^{(n)},
\boldsymbol{z}_1^{(n)}
\right)
\right]_{\varepsilon}
=
\mathbf{0}.
\label{eq:appendix-f2p-zero-epsilon}
\end{equation}
Consequently, under this exact pairing, the Monte Carlo guided vector field in \eqref{eq:appendix-enff-mc-guidance} also has a zero parameter component.

In LD-EnFF, we instead use a perturbed F2P coupling. Motivated by the Gaussian smoothing of the F2P conditional path at $t=0$,
\begin{equation}
p_0(\boldsymbol{z}\mid\boldsymbol{z}_0,\boldsymbol{z}_1)
=
\mathcal{N}
\!\left(
\boldsymbol{z};
\boldsymbol{z}_0,
\sigma_{\min}^2\mathbf{I}
\right),
\end{equation}
we draw
\begin{equation}
\widetilde{\boldsymbol{z}}_0^{(n)}
=
\boldsymbol{z}_0^{(n)}
+
\sigma_{\min}\boldsymbol{\xi}^{(n)},
\qquad
\boldsymbol{\xi}^{(n)}
\sim
\mathcal{N}(\mathbf{0},\mathbf{I}),
\label{eq:appendix-f2p-perturbation}
\end{equation}
and use $(\widetilde{\boldsymbol{z}}_0^{(n)},\boldsymbol{z}_1^{(n)})$ as the effective pair in the subsequent F2P and Monte Carlo guidance calculations. The resulting parameter component of the conditional velocity is
\begin{equation}
\left[
\boldsymbol{u}_t
\left(
\boldsymbol{z}\mid
\widetilde{\boldsymbol{z}}_0^{(n)},
\boldsymbol{z}_1^{(n)}
\right)
\right]_{\varepsilon}
=
-\sigma_{\min}\boldsymbol{\xi}_{\varepsilon}^{(n)},
\label{eq:appendix-f2p-perturbed-epsilon}
\end{equation}
which is generally nonzero and has characteristic magnitude $O(\sigma_{\min})$.

The perturbation is zero-mean and therefore does not by itself introduce a systematic drift in the uncertain parameter. Instead, it provides nonzero directions in the parameter coordinates that are subsequently reweighted through the observation-dependent Monte Carlo guidance. This allows the analysis to update $\boldsymbol{\varepsilon}$ while avoiding the zero-velocity degeneracy of the exact dynamical F2P coupling. By contrast, the OT construction starts from a fixed Gaussian reference and does not impose the same endpoint-matching constraint on the parameter coordinates.

\subsection{LD-EnFF Assimilation Algorithm}
\label{app:ld-enff-algorithm}

Algorithm~\ref{alg:ld-enff} summarizes one assimilation step of LD-EnFF.
\begin{algorithm}[H]
\caption{One assimilation step of LD-EnFF}
\label{alg:ld-enff}
\begin{algorithmic}[1]
\Statex\hspace{-\algorithmicindent}\textbf{Input:}
Posterior augmented latent ensemble
$\{\boldsymbol{\kappa}_{j-1}^{(n)}\}_{n=1}^{N}$;
physical observation $\mathbf{y}_j$;
time increment $\Delta t$;
observation-noise covariance $\boldsymbol{\Gamma}_y$;
trained $F_{\theta_1}$, $E_{\theta_3}$, and $H_\alpha$

\Statex\hspace{-\algorithmicindent}\textbf{Output:}
Posterior augmented latent ensemble
$\{\boldsymbol{\kappa}_{j}^{(n)}\}_{n=1}^{N}$

\For{$n=1,\ldots,N$}
    \State $\widehat{\mathbf{s}}_j^{(n)}
    \gets
    \mathbf{s}_{j-1}^{(n)}
    +
    \Delta t\,
    F_{\theta_1}
    \!\left(
    \mathbf{s}_{j-1}^{(n)},
    \boldsymbol{\varepsilon}_{j-1}^{(n)}
    \right)$
    \State $\widehat{\boldsymbol{\varepsilon}}_j^{(n)}
    \gets
    \boldsymbol{\varepsilon}_{j-1}^{(n)}$
    \State $\widehat{\boldsymbol{\kappa}}_j^{(n)}
    \gets
    (
    \widehat{\mathbf{s}}_j^{(n)},
    \widehat{\boldsymbol{\varepsilon}}_j^{(n)}
    )$
\EndFor

\State $\bar{\boldsymbol{\phi}}_j
\gets
\boldsymbol{\mu}_{\theta_3}(\mathbf{y}_j)$

\State Construct
$\mathcal{L}^{\mathrm{lat}}_j(\boldsymbol{\kappa})$
according to \eqref{eq:latent-likelihood}

\State Apply EnFF to
$\{\widehat{\boldsymbol{\kappa}}_j^{(n)}\}_{n=1}^{N}$
using $\mathcal{L}^{\mathrm{lat}}_j$ as the observation factor

\State Obtain
$\{\boldsymbol{\kappa}_j^{(n)}\}_{n=1}^{N}$
\end{algorithmic}
\end{algorithm}

\section{Benchmark Details}
\label{app:benchmark-details}

\subsection{Kolmogorov Flow}
\label{app:kolmogorov-details}

We consider two-dimensional Kolmogorov flow, a canonical turbulent system driven by a sinusoidal body force. Let $\mathbf{v}=(v_1,v_2)$ denote the velocity field and $p$ the pressure. On the periodic domain $[0,2\pi]^2$, the dynamics satisfy the incompressible Navier--Stokes equations \citep{kochkov2021machine}
\begin{equation}
\begin{aligned}
\frac{\partial\mathbf{v}}{\partial t}
+
(\mathbf{v}\cdot\nabla)\mathbf{v}
&=
-\nabla p
+
\frac{1}{\mathrm{Re}}\nabla^2\mathbf{v}
+
\mathbf{f},\\
\nabla\cdot\mathbf{v}
&=
0,
\end{aligned}
\label{eq:kolmogorov-pde}
\end{equation}
where, assuming unit fluid density,
\begin{equation}
\mathbf{f}
=
\sin(4\xi_2)\widehat{\boldsymbol{\xi}}_1
-
0.1\mathbf{v},
\label{eq:kolmogorov-forcing}
\end{equation}
with $\boldsymbol{\xi}=(\xi_1,\xi_2)$ denoting the spatial coordinate and
$\widehat{\boldsymbol{\xi}}_1=(1,0)$ the unit vector in the $\xi_1$ direction.
The Reynolds number $\mathrm{Re}$ is treated as the uncertain parameter and is sampled from $[500,1500]$.

We generate 200 trajectories on a $150\times150$ spatial grid. All trajectories start from the same fixed initial velocity field and differ through their sampled Reynolds numbers. Each trajectory is advanced for a 100-step burn-in period, followed by a 200-step assimilation window with simulation time step $\delta t=0.04$. Observations are assimilated every five simulation steps during the post-burn-in window, yielding 40 assimilation times. At each assimilation time, both velocity components are observed at 100 spatial locations on a fixed regular $10\times10$ grid, corresponding to $0.44\%$ of the $150\times150$ spatial locations and yielding 200 scalar observations.

\subsection{Tsunami Modeling}
\label{app:tsunami-details}

We consider two-dimensional shallow-water wave propagation for tsunami modeling, following the setup in \citet{si2025latentensf}. Let $\mathbf{v}=(u,v)$ denote the horizontal velocity field and $\eta$ the free-surface elevation. The dynamics are governed by the simplified non-rotating shallow-water equations
\begin{equation}
\begin{aligned}
\frac{\partial\mathbf{v}}{\partial t}
&=
-g\nabla\eta,\\
\frac{\partial\eta}{\partial t}
&=
-\nabla\cdot\bigl((H+\eta)\mathbf{v}\bigr),
\end{aligned}
\label{eq:shallow-water-pde}
\end{equation}
where $g=9.81\,\mathrm{m/s^2}$ is the gravitational acceleration and
$H=100\,\mathrm{m}$ is the mean water depth. The assimilated state consists of the two horizontal velocity components and the free-surface elevation.

The spatial domain is a square of side length $L=10^6\,\mathrm{m}$. The initial velocity is set to $\mathbf{v}=\mathbf{0}$, while the initial surface elevation is prescribed as a unit-amplitude Gaussian bump with standard deviation $0.05\times10^6\,\mathrm{m}$. The center of this Gaussian perturbation is sampled uniformly over the spatial domain for each trajectory and serves as the uncertain parameter.

We generate 200 trajectories on a uniform $150\times150$ spatial grid. Each trajectory is integrated for 2000 solver steps using an upwind discretization with a CFL-based time step of approximately $21$ seconds. States are recorded every 40 solver steps. Including the initial condition, this produces 51 stored frames and 50 subsequent assimilation times. At each assimilation time, all three state components are observed at 100 locations on a fixed regular $10\times10$ grid, corresponding to $0.44\%$ of the spatial locations and yielding 300 scalar observations.

\subsection{Atmospheric Modeling}
\label{app:atmospheric-details}

We consider PlanetSWE, a rotating shallow-water model on the sphere for atmospheric modeling, following the setup in \citet{ohana2024well}. Let $\mathbf{v}$ denote the tangential velocity field, $h$ the deviation of the pressure-surface height from its mean value $H$, and $h_s$ the prescribed topographic height. Using spherical differential operators, the governing equations are
\begin{equation}
\begin{aligned}
\frac{\partial\mathbf{v}}{\partial t}
&=
-(\mathbf{v}\cdot\nabla)\mathbf{v}
-g\nabla h
-\nu\nabla^4\mathbf{v}
-2\boldsymbol{\Omega}\times\mathbf{v},\\
\frac{\partial h}{\partial t}
&=
-H\nabla\cdot\mathbf{v}
-\nabla\cdot\bigl(\mathbf{v}(h-h_s)\bigr)
-\nu\nabla^4 h
+F(\phi,\theta,t),
\end{aligned}
\label{eq:planetswe-pde}
\end{equation}
where $g$ is the gravitational acceleration, $\boldsymbol{\Omega}$ is the planetary rotation vector, and $\nu$ is the hyperviscosity coefficient. The initial velocity and height fields are derived from ERA5 reanalysis data at the 500-hPa pressure level.

The external forcing has a longitudinal center that varies with the daily cycle and a latitudinal center that varies with the seasonal cycle:
\begin{equation}
\begin{aligned}
\phi_c(t)
&=
2\pi\frac{t}{\mathrm{day}},\\
\theta_c(t)
&=
\theta_{\max}
\sin\!\left(
2\pi\frac{t}{\mathrm{year}}
\right),\\
F(\phi,\theta,t)
&=
h_f^0
\cos\bigl(\phi-\phi_c(t)\bigr)
\exp\!\left(
-\frac{(\theta-\theta_c(t))^2}{\sigma^2}
\right),
\end{aligned}
\label{eq:planetswe-forcing}
\end{equation}
where $\phi$ and $\theta$ denote longitude and latitude, respectively,
$\theta_{\max}=0.4$, $h_f^0$ is the forcing amplitude, and $\sigma$ controls the latitudinal spread of the forcing. The model uses a 42-day year. For each trajectory, the uncertain forcing parameters are sampled as
$h_f^0\sim\mathcal{U}[0.1,30]$
and
$\sigma\sim\mathcal{U}[1,4]$.

We generate 200 trajectories on a $512\times256$ spherical grid. Each trajectory is simulated for 21 days using adaptive CFL-controlled time integration, with an internal time step of approximately $60$ seconds. Observations are assimilated every eight hours, yielding 63 assimilation times. At each assimilation time, all three state components are observed at 128 locations on a regular $16\times8$ grid, corresponding to approximately $0.1\%$ of the $512\times256$ spatial locations and yielding 384 scalar observations. The resulting physical-state dimension is
$512\times256\times3=393{,}216$.

\section{Training and Implementation Details}
\label{app:training-details}

\subsection{System Configuration}
\label{app:system-configuration}

All models were implemented in PyTorch, with DeepXDE used for LDNet training. Experiments used Python 3.10.20, PyTorch 2.4.1, and CUDA 12.1. For all three benchmarks, each LDNet, variational observation encoder, and $y^\star$ operator training run used a single NVIDIA V100 GPU, 8 CPU cores, and 32 GB of system memory. All training and evaluation runs were performed on a single GPU, with no distributed training.

\subsection{Hyperparameter Choice}
\label{app:hyperparameter-choice}

\noindent\textbf{Normalization.}
Observation noise is added in the original physical observation space before applying the benchmark-specific normalization. Accordingly, when evaluating the encoder-based latent-observation likelihood, the observation-noise covariance is transformed under the same normalization. For Kolmogorov flow, the two velocity components are standardized using the mean and standard deviation computed from the training split, while the Reynolds number is linearly mapped from $[500,1500]$ to $[-1,1]$. For shallow water, the two velocity components and the free-surface elevation are each linearly mapped to $[-1,1]$ using the fixed ranges $[-0.15,0.15]$, $[-0.15,0.15]$, and $[-0.5,1.0]$, respectively, while the two-dimensional initial-disturbance location is normalized to $[-1,1]^2$. For PlanetSWE, the height and two velocity components are standardized using statistics computed from the training split, while the forcing parameters $(h_f^0,\sigma)$ are linearly mapped from $[0.1,30]\times[1,4]$ to $[-1,1]^2$.

\noindent\textbf{LDNet.}
For all three benchmarks, we adopt the applicable LDNet training hyperparameters from the LD-EnSF setup \citep{xiao2026ldensf}. The hyperparameters used in our experiments are summarized in Table~\ref{tab:ldnet-training-details}. ``Downsampled time steps'' denotes the number of temporal snapshots retained from each trajectory for LDNet training. The $\Delta t$ values in Table~\ref{tab:ldnet-training-details} denote the normalized model-time integration increments used by LDNet. The dynamics and reconstruction networks use MLPs with ReLU activations and Glorot-normal initialization. LDNet is trained for $2000$ epochs using Adam with a learning rate of $10^{-3}$ and the reconstruction loss in \eqref{eq:ldnet-loss}. A StepLR scheduler with step size $200$ is used, with the benchmark-specific decay factors reported in Table~\ref{tab:ldnet-training-details}. After the initial LDNet training, the reconstruction network is further fine-tuned for $1000$ epochs using a learning rate of $10^{-4}$.

\begin{table*}[!htbp]
\centering
\caption{LDNet training hyperparameters used for the three benchmarks.}
\label{tab:ldnet-training-details}
\begin{tabular}{lccc}
\toprule
Configuration & Kolmogorov Flow & Shallow Water & PlanetSWE \\
\midrule
Downsampled time steps & $40$ & $50$ & $63$ \\
$\Delta t$ & $0.04$ & $0.036$ & $0.04$ \\
Latent-state dimension & $9$ & $10$ & $50$ \\
Dynamics network hidden layers & $9$ & $8$ & $8$ \\
Dynamics network hidden dimension & $200$ & $50$ & $200$ \\
Reconstruction network hidden layers & $14$ & $10$ & $15$ \\
Reconstruction network hidden dimension & $500$ & $300$ & $500$ \\
StepLR decay factor $\gamma$ & $0.7$ & $0.6$ & $0.6$ \\
Batch size & $6$ & $2$ & $2$ \\
\bottomrule
\end{tabular}
\end{table*}

\noindent\textbf{Variational observation encoder.}
For each benchmark, the variational observation encoder takes the sparse physical observation arranged on its observation grid and parameterizes a diagonal Gaussian distribution in the latent observation space. The encoder outputs the mean and elementwise log-variance, each represented with four latent channels. We use the same compact residual VAE architecture for all three benchmarks, with hidden dimension $8$ and one residual block. The observation inputs have shapes $2\times10\times10$, $3\times10\times10$, and $3\times16\times8$ for Kolmogorov flow, shallow water, and PlanetSWE, respectively.

During training, the observation decoder $O_{\theta_4}$ maps the sampled latent variable to $\widehat{\mathbf{y}}_j=O_{\theta_4}(\boldsymbol{\phi}_j)$, with the measured noisy observation $\mathbf{y}_j$ as the reconstruction target. For $N_{\mathrm{tr}}$ training trajectories, the encoder and decoder parameters are learned by minimizing
\begin{equation}
\begin{aligned}
\mathcal{L}_{\mathrm{VAE}}(\theta_3,\theta_4)
=
\frac{1}{N_{\mathrm{tr}}J}
\sum_{i=1}^{N_{\mathrm{tr}}}
\sum_{j=1}^{J}
\Bigg[
&
\mathbb{E}_{q_{\theta_3}(\boldsymbol{\phi}\mid\mathbf{y}_j^{(i)})}
\left[
\left\|
O_{\theta_4}(\boldsymbol{\phi})
-
\mathbf{y}_j^{(i)}
\right\|_2^2
\right]
\\
&+
\beta
D_{\mathrm{KL}}\!\left(
q_{\theta_3}(\boldsymbol{\phi}\mid\mathbf{y}_j^{(i)})
\,\|\,
\mathcal{N}(\mathbf{0},\mathbf{I}_{d_\phi})
\right)
\Bigg].
\end{aligned}
\label{eq:vae-loss}
\end{equation}
Here, $\beta>0$ controls the strength of the KL regularization.

The model is trained for $400$ epochs using Adam with a learning rate of $10^{-3}$, batch size $100$, and KL weight $\beta=10^{-5}$. The learning rate is reduced by a factor of $0.1$ at epochs $100$, $200$, $250$, and $300$. The checkpoint with the lowest validation loss is retained.

\noindent\textbf{$y^\star$ operator.}
For each benchmark, the $y^\star$ operator is trained after the variational observation encoder. Its input is the augmented latent state $\boldsymbol{\kappa}=(\mathbf{s},\boldsymbol{\varepsilon})$, and its training target is the corresponding noiseless sparse physical observation. All three operators are trained for $400$ epochs by minimizing the squared-error objective in \eqref{eq:ystar-loss} using Adam with an initial learning rate of $10^{-3}$. The checkpoint with the lowest validation loss is retained. The dataset-specific input dimensions, output shapes, MLP hidden dimensions, and batch sizes are summarized in Table~\ref{tab:ystar-training-details}.

\begin{table*}[!htbp]
\centering
\caption{Learned $y^\star$ operator configurations for the three benchmarks.}
\label{tab:ystar-training-details}
\small
\setlength{\tabcolsep}{4pt}
\begin{tabular}{@{}lccc@{}}
\toprule
Configuration & Kolmogorov Flow & Shallow Water & PlanetSWE \\
\midrule
Augmented latent input dimension & $10$ & $12$ & $52$ \\
Output shape & $2\times10\times10$ & $3\times10\times10$ & $3\times16\times8$ \\
MLP hidden dimensions & $(128,256,512,1024)$ & $(256,512,1024,2048)$ & $(256,512,1024,2048)$ \\
Batch size & $100$ & $100$ & $100$ \\
\bottomrule
\end{tabular}
\end{table*}

\subsection{Compared Methods}
\label{app:compared-methods}

Table~\ref{tab:method-summary} summarizes the methods used in the data-assimilation comparisons and their main components.

\begin{table*}[!htbp]
\centering
\caption{Summary of the methods compared in the experiments.}
\label{tab:method-summary}
\small
\setlength{\tabcolsep}{4pt}
\begin{tabular}{@{}llll@{}}
\toprule
Method & State propagation & Observation construction & Analysis \\
\midrule
LD-EnFF-OT (ours)      & LDNet      & VAE encoder + $y^\star$ operator & EnFF-OT \\
LD-EnFF-F2P (ours)     & LDNet      & VAE encoder + $y^\star$ operator & EnFF-F2P \\
LD-IEnKF               & LDNet      & VAE encoder + $y^\star$ operator & IEnKF \\
LDNet+LSTM EnSF        & LDNet      & LSTM encoder & EnSF \\
LDNet+LSTM EnFF-OT     & LDNet      & LSTM encoder & EnFF-OT \\
LDNet+LSTM EnFF-F2P    & LDNet      & LSTM encoder & EnFF-F2P \\
LDNet dynamics only    & LDNet      & -- & None \\
LETKF                   & Full model & Physical observation & LETKF \\
Full-space EnFF-OT     & Full model & Physical observation & EnFF-OT \\
Full-space EnFF-F2P    & Full model & Physical observation & EnFF-F2P \\
\bottomrule
\end{tabular}
\end{table*}

\subsection{Evaluation Protocol Details}
\label{app:evaluation-protocol}

For each benchmark, the 200 generated trajectories are partitioned into $60\%$ for training, $20\%$ for validation, and $20\%$ for testing. The learned components are fitted using the training trajectories, while the validation trajectories are used for architecture selection and hyperparameter tuning. The test trajectories are held out from both training and validation.

For the reported data-assimilation comparisons, all methods within a given benchmark are evaluated on the same randomly selected subset of 20 trajectories from the 40 held-out test trajectories, with three random seeds used for each trajectory. For each trajectory and seed, the corresponding sparse observation sequence is generated using the prescribed observation operator and additive Gaussian observation noise with covariance $\boldsymbol{\Gamma}_y=0.1^2\mathbf{I}_{d_y}$. All compared methods use the same ground-truth trajectory and the realized sparse noisy observation sequence for a given run.

All ensemble-based methods use $N=20$ ensemble members. The true uncertain parameter associated with each test trajectory is not used in constructing the initial ensemble. For each trajectory and seed, all latent-space methods are initialized from the same augmented latent ensemble. For the full-space baselines, the initial augmented physical-state ensemble is constructed using the same parameter samples as those used for the latent-space methods. For Kolmogorov Flow and Atmospheric Modeling, the physical-state components are identical across ensemble members at initialization, while the parameter components differ; forecast propagation then induces physical-state ensemble spread because the dynamics depend on the uncertain parameter. For Tsunami Modeling, the uncertain parameter specifies the center of the initial Gaussian surface perturbation, so different parameter samples directly produce different initial physical states. Thus, in all three benchmarks, the full-space baselines are initialized consistently with the benchmark-specific role of the uncertain parameter without access to its true value.

\section{Uncertain-Parameter Error and Additional Assimilation Visualizations}
\label{app:visualization}

This section provides two types of additional results for all three benchmarks. First, we report uncertain-parameter relative-error trajectories for all assimilating latent-space methods to examine how the uncertain parameter is updated during sequential assimilation. Second, we provide final-state visualizations showing representative physical fields and their pointwise errors at the final assimilation time. Together, these results provide additional diagnostic and qualitative evidence for the assimilation performance reported in Section~\ref{sec:assimilation-results}.

\subsection{Uncertain-Parameter Relative Error}
\label{app:uncertain-parameter-error}

To evaluate the estimation of the time-invariant uncertain parameter, we compute the relative error of its posterior ensemble mean, $\widehat{\boldsymbol{\varepsilon}}_j=N^{-1}\sum_{n=1}^{N}\boldsymbol{\varepsilon}_j^{(n)}$, as
\begin{equation}
E_{\varepsilon,j}(\%)
=
100\,
\frac{
\left\|
\widehat{\boldsymbol{\varepsilon}}_j-\boldsymbol{\varepsilon}^{\star}
\right\|_2
}{
\left\|
\boldsymbol{\varepsilon}^{\star}
\right\|_2
},
\label{eq:uncertain-parameter-relative-error}
\end{equation}
where $\boldsymbol{\varepsilon}^{\star}$ denotes the true uncertain parameter associated with a given test trajectory. The relative error is computed after mapping both the estimated and true uncertain parameters back to their original physical coordinates.

\begin{figure*}[!htbp]
\centering
\includegraphics[width=\textwidth]{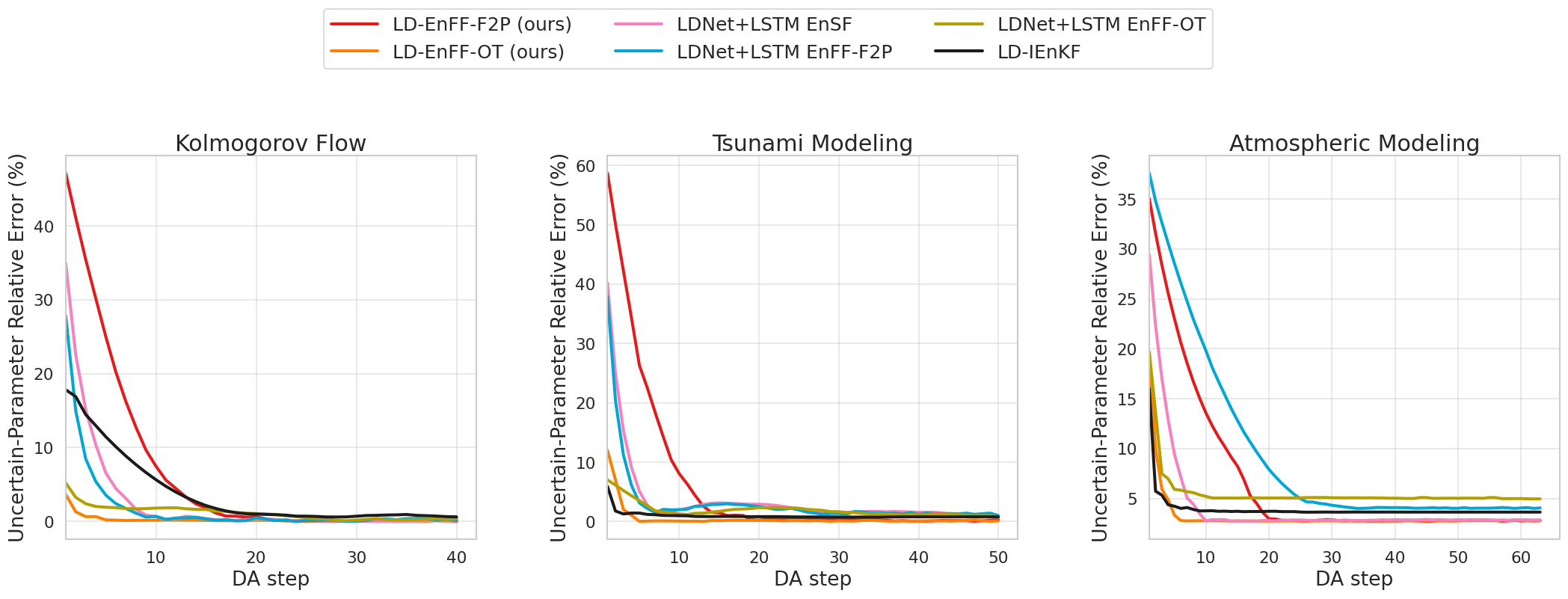}
\caption{Mean uncertain-parameter relative error over 20 held-out test trajectories and three random seeds per trajectory for all assimilating latent-space methods on Kolmogorov Flow (left), Tsunami Modeling (center), and Atmospheric Modeling (right).}
\label{fig:parameter-error}
\end{figure*}

Figure~\ref{fig:parameter-error} compares the uncertain-parameter error for all assimilating latent-space methods. Across the three benchmarks, all methods substantially reduce the error from its initial level, showing that the uncertain parameter is updated through sequential assimilation rather than remaining fixed at initialization. Within LD-EnFF, OT generally reduces the uncertain-parameter error more rapidly than F2P during the early assimilation stages, consistent with the full-state error behavior in Figure~\ref{fig:all-datasets-results}. The more gradual F2P adjustment is consistent with the perturbed F2P mechanism described in Appendix~\ref{app:f2p-perturbation}. The no-assimilation LDNet reference is omitted because it does not update the uncertain parameter through an analysis step.

\subsection{Final-State Visualizations}

\begin{figure*}[!htbp]
\centering
\includegraphics[
    width=\textwidth,
    trim=70bp 0 0 0,
    clip
]{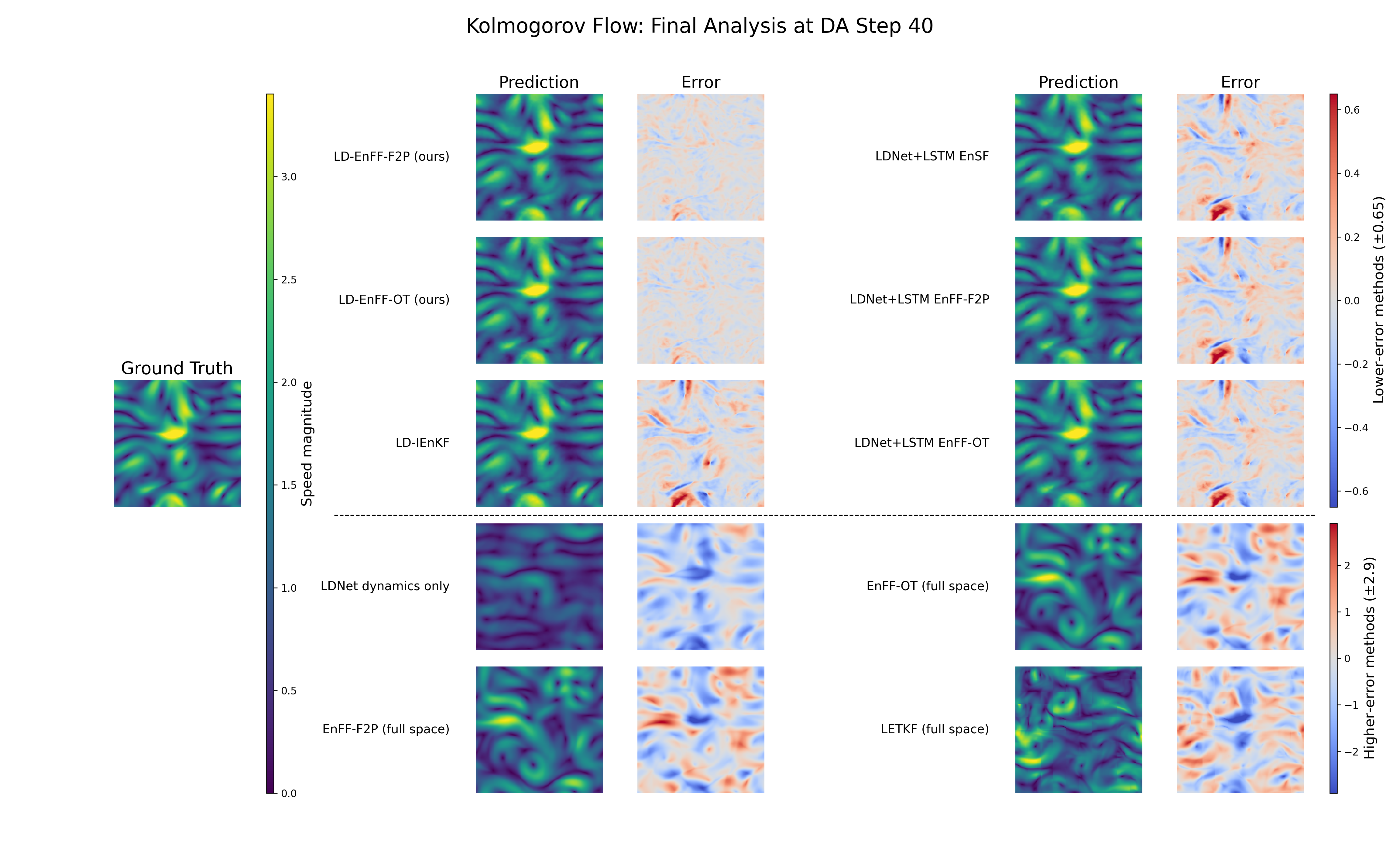}
\caption{Speed-magnitude fields and pointwise errors at the final assimilation time for Kolmogorov Flow. The ground truth is shown on the left, followed by the estimates and corresponding errors for the compared methods. Each pointwise error is defined as the prediction minus the ground truth.}
\label{fig:kolmogorov-speed}
\end{figure*}

\begin{figure*}[!htbp]
\centering
\includegraphics[
    width=\textwidth,
    trim=70bp 0 0 0,
    clip
]{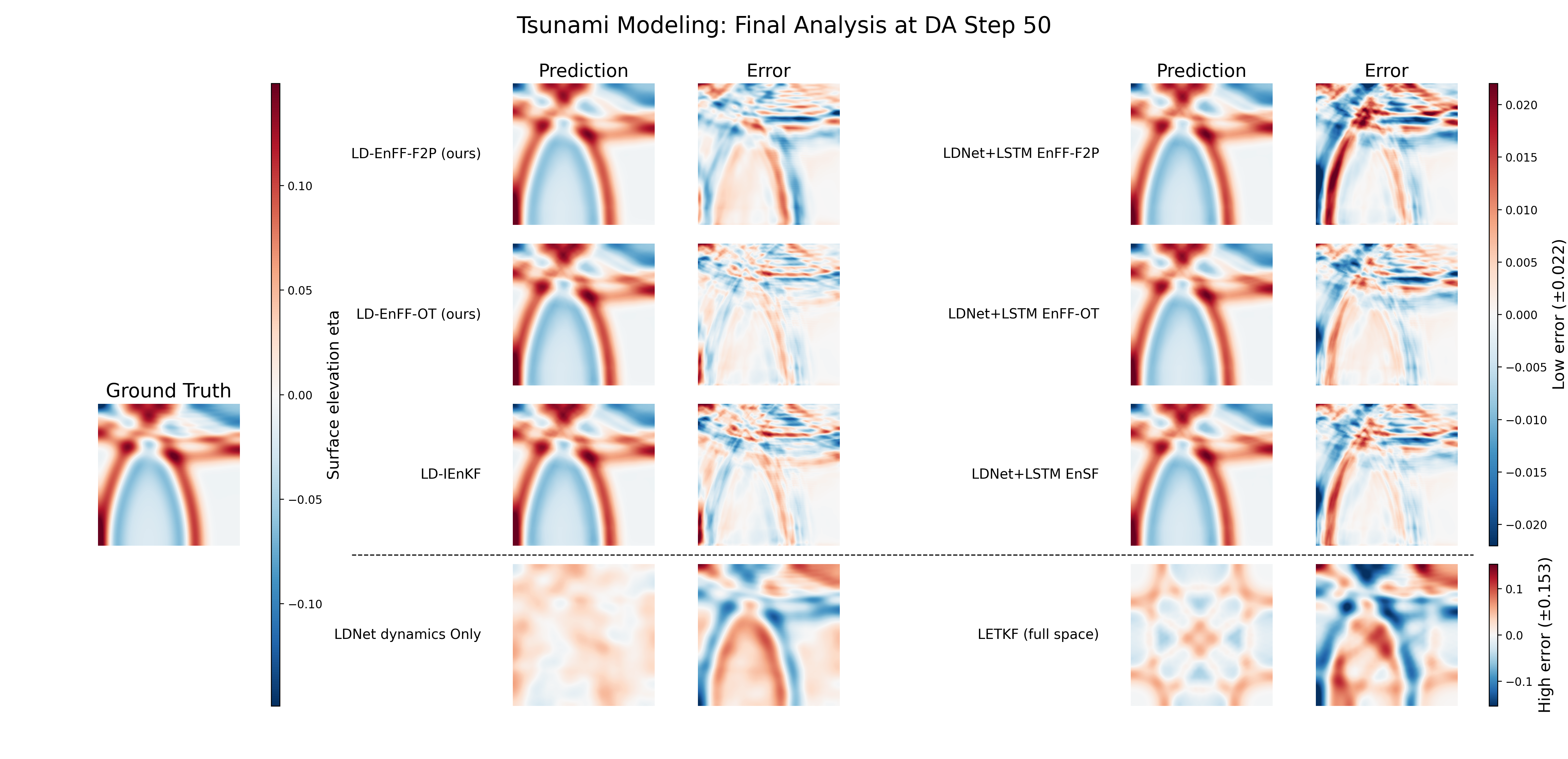}
\caption{Free-surface elevation fields and pointwise errors at the final assimilation time for Tsunami Modeling. The ground truth is shown on the left. For each compared method, the prediction and corresponding pointwise error are shown side by side. Each pointwise error is defined as the prediction minus the ground truth.}
\label{fig:shallow-water-final}
\end{figure*}

\begin{figure*}[!htbp]
\centering
\includegraphics[width=\textwidth]{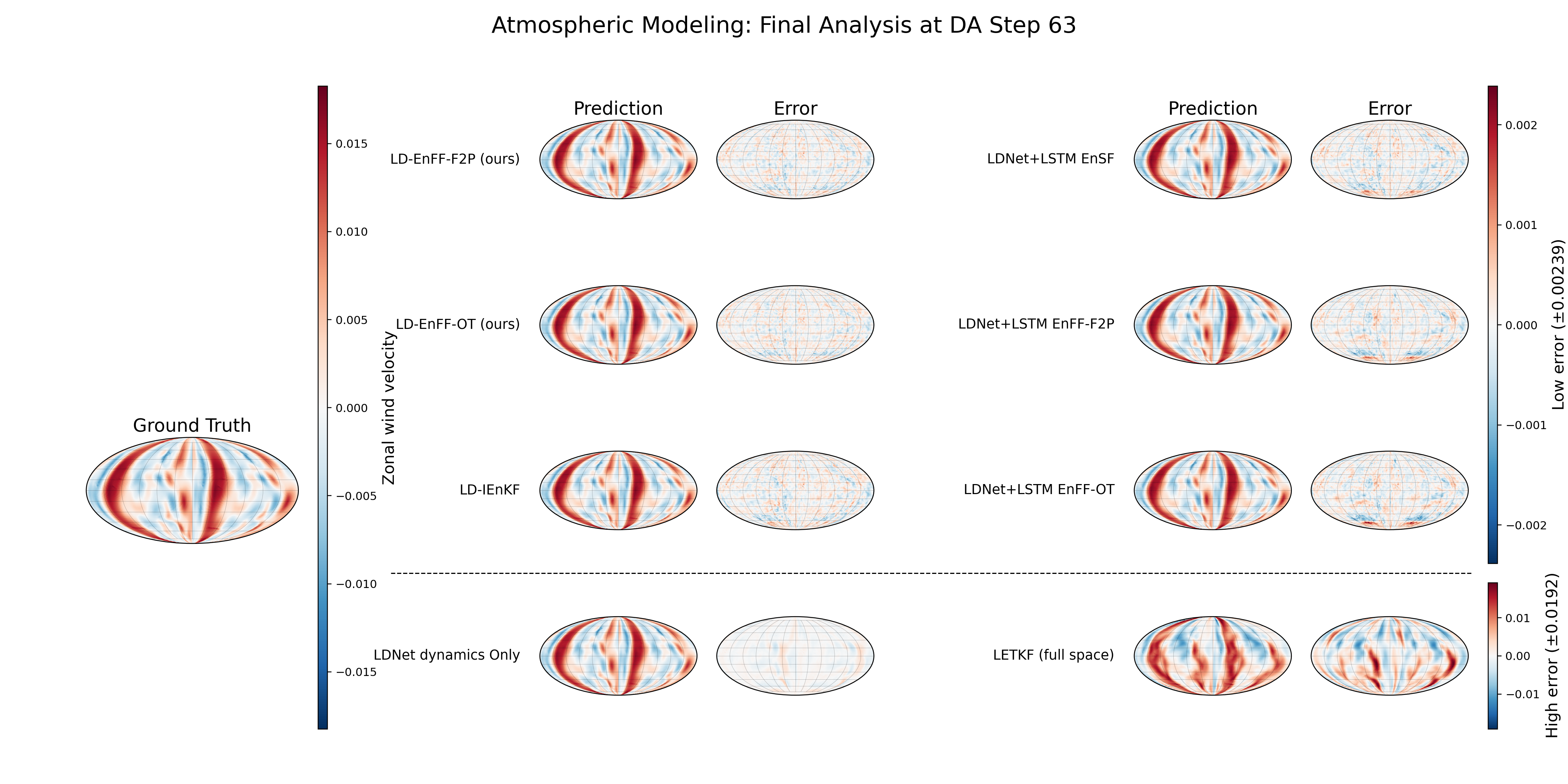}
\caption{Zonal wind velocity fields and pointwise errors at the final assimilation time for Atmospheric Modeling. The ground truth is shown on the left. For each compared method, the estimate and corresponding pointwise error (estimate minus ground truth) are shown side by side.}
\label{fig:planetswe-final}
\end{figure*}

Final-state visualizations for the three benchmarks are provided in Figures~\ref{fig:kolmogorov-speed}, \ref{fig:shallow-water-final}, and \ref{fig:planetswe-final}. These figures show representative physical fields together with their pointwise errors at the final assimilation time.

\end{document}